\documentclass[journal]{IEEEtran}

\usepackage{xcolor,soul,framed} 
\usepackage{booktabs} 
\usepackage{multirow}
\usepackage{longtable}
\usepackage{graphicx}
\usepackage{array}
\usepackage{tabulary}
\usepackage{lscape}
\usepackage{tabu}
\usepackage{adjustbox}
\usepackage{comment}

\colorlet{shadecolor}{yellow}
\usepackage{graphicx}

\usepackage{array}
\usepackage{mdwmath}
\usepackage{mdwtab}
\usepackage{eqparbox}
\usepackage{url}

\usepackage{color}

\usepackage{soul}

\begin{document}
\bstctlcite{IEEEexample:BSTcontrol}
    \title{Different Approaches for Human Activity Recognition-- A Survey}
  \author{Zawar~Hussain,~
      Michael~Sheng,~\IEEEmembership{Senior Member,~IEEE,}\\
      Wei Emma~Zhang,~\IEEEmembership{Member,~IEEE}

 \thanks{Z. Hussain is a PhD scholar in Department of Computing at Macquarie University, Sydney, Australia (e-mail: zawar.hussain@hdr.mq.edu.au).}
 \thanks{M. Sheng is a full Professor and Head of the Computing Department at Macquarie University, Sydney, Australia (e-mail: michael.sheng@mq.edu.au).}%
 \thanks{W. Zhang is a senior Post Doctoral Research Fellow in Department of Computing at Macquarie University, Sydney, Australia (e-mail: w.zhang@mq.edu.au).}
}


\maketitle

\begin{abstract}
Human activity recognition has gained importance in recent years due to its applications in various fields such as health, security and surveillance, entertainment, and intelligent environments. A significant amount of work has been done on human activity recognition and researchers have leveraged different approaches, such as wearable, object-tagged, and device-free, to recognize human activities. In this article, we present a comprehensive survey of the work conducted over the period 2010-2018 in various areas of human activity recognition with main focus on device-free solutions. The device-free approach is becoming very popular due to the fact that the subject is not required to carry anything, instead, the environment is tagged with devices to capture the required information. We propose a new taxonomy for categorizing the research work conducted in the field of activity recognition and divide the existing literature into three sub-areas: action-based, motion-based, and interaction-based. We further divide these areas into ten different sub-topics and present the latest research work in these sub-topics. Unlike previous surveys which focus only on one type of activities, to the best of our knowledge, we cover all the sub-areas in activity recognition and provide a comparison of the latest research work in these sub-areas. Specifically, we discuss the key attributes and design approaches for the work presented. Then we provide extensive analysis based on 10 important metrics, to give the reader, a complete overview of the state-of-the-art techniques and trends in different sub-areas of human activity recognition. 
In the end, we discuss open research issues and provide future research directions in the field of human activity recognition.
\end{abstract}


\begin{IEEEkeywords}
human activity recognition, device-free, dense sensing, RFID
\end{IEEEkeywords}

\IEEEpeerreviewmaketitle

\section{Introduction}

\IEEEPARstart{H}activity recognition (HAR) has been a very active research topic for the past two decades for its applications in various fields such as health, remote monitoring, gaming, security and surveillance, and human-computer interaction. Activity recognition can be defined as the ability to recognize/detect current activity on the basis of information received from different sensors \cite{yang2011activity}. These sensors can be cameras, wearable sensors, sensors attached to objects of the daily use or deployed in the environment.
With the advancement in technology and reduction in devices cost, logging of daily activities has become very popular and practical. People are logging their daily life activities, such as preparing a meal, eating, sleeping, watching a TV, or the number of steps taken. To capture these activities, different approaches have been used. These approaches can be broadly classified into vision-based and sensor-based \cite{RN79} as shown in Figure 1. One of the pioneer approaches in this area is vision-based approach, in which a camera is used to capture the information about the activities of human. By applying computer vision techniques on this captured data, different activities can be recognized. Although computer vision based techniques are easy to use and can provide good results, there are many issues related to this approach. Privacy is the main concern. Another issue with this approach is light dependency. Traditional cameras fail to work if there is no proper light (e.g., night time). Many surveys have been written for vision-based approaches because it was one of the initial approaches used for activity recognition \cite{aggarwal2014human}. Therefore, vision-based techniques are not included in our survey.

\begin{figure}[hb]
	\centering
	\includegraphics[width=  \linewidth]{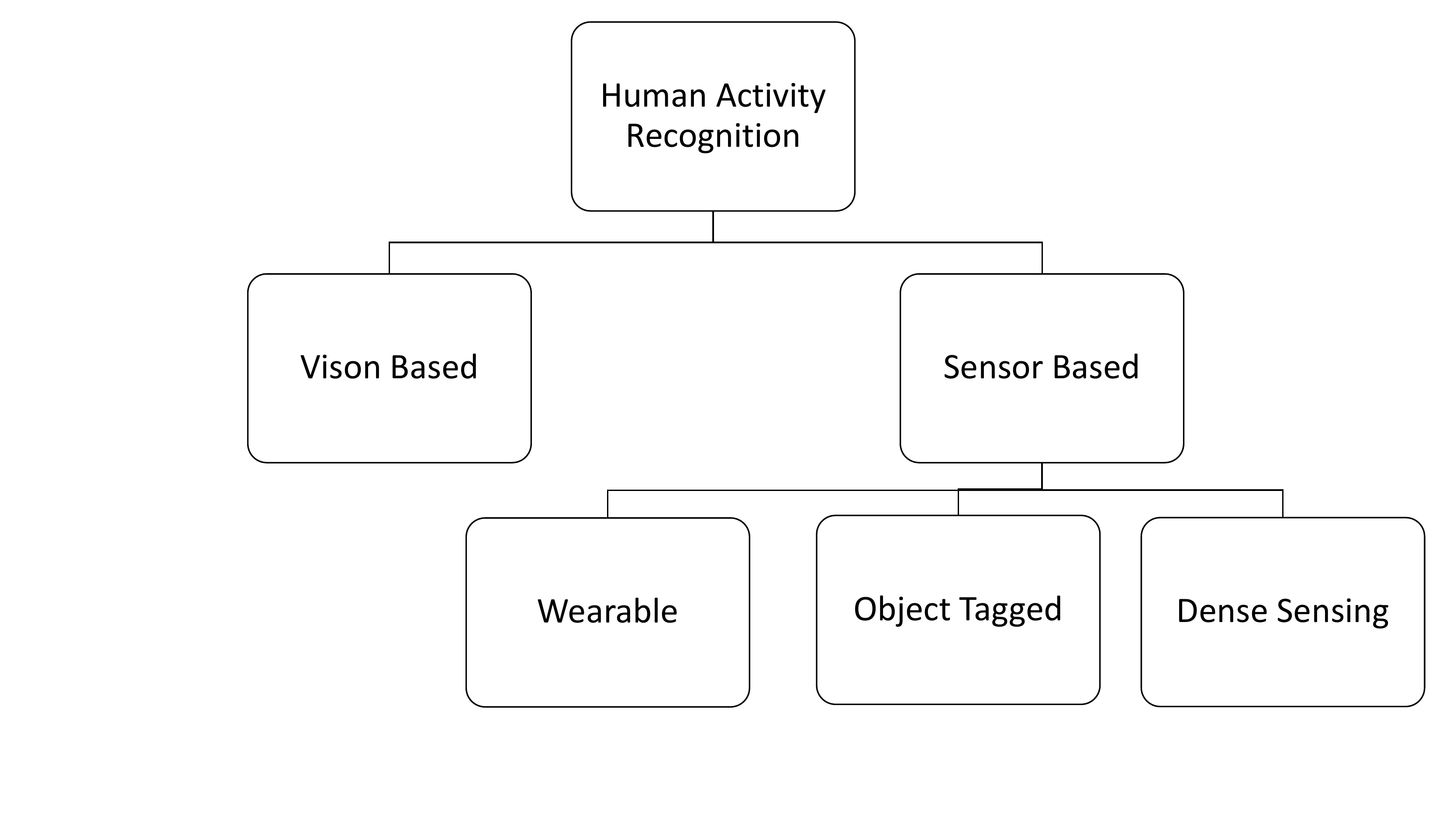}
	\caption{Classification of human activity recognition approaches}
	\label{fig1}
\end{figure}

Due to low cost and advancement in sensor technology, most of the research in the field of HAR has shifted towards a sensor-based approach. In the sensor-based approach, different sensors are used to capture the behavior of human while they perform daily life activities. Sensor-based solutions can further be divided into three major categories on the basis of sensor's deployment, which are: i) wearable, ii) object-tagged (device-bound), and iii) dense sensing (environment tagged/device-free) \cite{RN83}. 
In wearable approach, a user has to carry the sensors with them as they perform any activity. A significant amount of work has been done on activity recognition using wearable sensors but the major problem with this kind of approach is that wearing a tag is not feasible sometimes. For example, in the case of elderly or patients, they may forget to wear the tags or maybe, they resist to wear the tags at all. 
For solutions which use object-tagged approach, sensors are attached to objects of daily use. Based on a user's interaction with these objects, different activities are recognized. This is a device-bound approach, i.e., users are required to use specific objects (tagged-objects) only. Like wearable approach, this approach may also not be feasible all the time because it bounds the users to use tagged-objects.

Over the past few years, researchers are focusing on device-free (dense sensing) approach in which users are not required to carry any tag or device with them. The idea is to deploy sensors in the environment (the facility in which the activity is being performed) and when a person performs any activity, the data will be captured through those sensors, which can then be used for activity recognition. The device-free approach is more practical because it does not require the user to carry any device while doing any activity. But there are some challenges in this approach as well such as interference from the environment. The data captured by the sensors can be disturbed from the surroundings which can cause noise in the data.

In this survey, we provide an overview of the research works conducted over the period 2010-2018 in the field of human activity recognition with a focus on device-free approaches, especially the ones based on Radio Frequency Identification (RFID) technology. We divide the existing literature in activity recognition into three main categories, which are: i) action-based, ii) motion-based, and iii) interaction-based activities. These categories are further divided into 10 sub-areas. Research works for action-based activities are divided into gesture recognition, posture recognition, fall detection, activities of daily living, behavior recognition, and ambient assisted living. Motion-based activities are divided into tracking, motion detection, and people counting. Research works for interaction-based activities are grouped in a single category which is human-object interaction. We present the latest research in all these sub-areas of human activity recognition. We discuss and analyze the latest work in these areas to give the reader a comprehensive overview of the current research trends in the field of human activity recognition. The rest of the paper is organized as follows. Section 2 presents some related work and Section 3 provides technical details about the different technologies used in human activity recognition. Section 4 provides different categories of HAR and details of the work conducted in each category. Section 5 presents applications of HAR and in Section 6, we provide future challenges and open research issues in HAR. Section 7 discusses the issues faced while reviewing the literature and section 8 concludes this work.

\section{Related Work}

A considerable amount of work has been done in human activity recognition for the last decade. There are many surveys which summarize the research work in the area of activity recognition. These surveys focus on different approaches used for activity recognition and can be broadly classified into four main categories which are given as follows.

\subsection{Radio Frequency Based}
Surveys in this category focus on radio frequency (RF) based approaches for human activity recognition. Some of these surveys are discussed in this section.

Scholz et al. \cite{RN78} presented a survey of the research work in the field of device-free radio based activity recognition. This survey categorizes the existing work in device-free radio-based localization (DFL) and device-free radio-based activity recognition (DFAR). For DFL, the authors provide a description of different topics such as accurate presence detection, spatial coverage, adaptive machine learning, radio tomographic, and statistical modeling. For DFAR, the literature is sub-divided as adaptive threshold-based DFAR, machine learning-based DFAR, and statistical modeling-based DFAR. This work also provides a discussion on open challenges in the field of activity recognition. This survey mainly focuses on the analysis of radio sensor's usability in activity recognition.

Amendola et al. \cite{RN81} presented a survey summarizing the use of RFID technology for the Internet of Things (IoT) based health-related applications. This work describes the various uses of RFID tags such as environmental passive sensors which include volatile compound sensor and temperature sensors, and body-centric tags which include wearable tags and implantable tags. This work also provides some applications of RFID technology in human behavior analysis such as tracking, gesture recognition, and remote monitoring. The authors provide research directions in the field of RFID technology. This work discusses the possible use of RFID technology in various applications but does not provide any details about the work done in those application areas.

Wang \& Zhou \cite{RN83} summarized research work in the field of radio-based activity recognition. This survey categorizes the existing work in four major categories: i) ZigBee radio-based, ii) Wi-Fi-based, iii) RFID-based, and iv) other radio-based (e.g., FM radio, microwave). The authors present a comparison of all these techniques using metrics like coverage, accuracy, activity types, and deployment costs. They also provide some future research directions. This work focuses on only a single device-free approach which is based on RFID.

Ma et al. \cite{RN86} provided a short survey of the research in activity recognition using Wi-Fi-based approach. The paper gives a brief overview of the key technologies in Wi-Fi related work from the literature, to formulate a framework for activity recognition system, based on Wi-Fi. The major steps for this framework are base signal selection, pre-processing, feature extraction, and classification techniques. Three different kinds of base signals are discussed which includes amplitude, phase, and phase difference. The step of pre-processing is sub-divided into outline removal, irrelevant information removal, and redundancy removal. Feature extraction step involves space transformation and feature selection. Finally, in the classification step, two methods are discussed which are rule-based and machine learning based. This survey categorizes the literature in activity recognition into two major groups: coarse-grained activities and fine-grained activities. This work discusses only Wi-Fi-based research and also the detail about the given work is missing. This survey lacks the comparison between different approaches discussed and instead focuses on the steps involved in a Wi-Fi-based human activity recognition model.

The survey presented by Cianca et al. \cite{RN87} outlines the work conducted in the field of human activity recognition using RF signals. The authors classify human activity recognition into sub-categories such as presence detection, fall detection, activity detection, gesture and posture recognition, people counting, personal characteristic identification, breathe and vital sign detection, and human-object interactions. This work is mainly focused on device-free passive sensing approaches and divides these approaches on the bases of signal characteristics (bandwidth, carrier frequency, and transmission mode), type of measurement on the received signal (directly generated CSI or raw data from SDR platform), and type of signal descriptor used. This survey paper provides a good outline of the activity recognition work using RF signals.

\subsection{Sensor Based}
This section presents the surveys which focus on sensor-based approaches for activity recognition.

Chen et al. \cite{RN79} presented a detailed survey of the sensor based work in human activity recognition. This survey classifies the existing research efforts in two main categories: i) vision-based vs sensor-based, and ii) data-driven based vs knowledge-driven based. In the first categorization, the survey focuses on sensor-based approaches. Different techniques are discussed which use wearable sensors (e.g., accelerometer, GPS, and biosensors) and dense sensing. In the second way of classification, authors categorize the literature in activity recognition into data-driven vs knowledge-driven approaches. For data-driven approaches, the authors discuss techniques using generative modeling and discriminative modeling. For knowledge-driven approaches, techniques are further divided into logic-based, ontology-based, and mining based approaches. The main focus of this survey is data-centric activity recognition techniques.

Another survey by Wang et al. \cite{RN90} highlighted the different deep learning approaches for human activity recognition, using sensors. This work classifies the literature in activity recognition on the basis of sensor modality, deep model, and application area. On the basis of modality, the literature is divided into four aspects: body-worn sensors, object sensors, ambient sensors, and hybrid sensors. On the basis of the deep model, related work is categorized as discriminative deep architecture, generative deep architecture, and hybrid deep architecture. With respect to the application area, the related work is classified as activities of daily living, sleep, sports, and health. This survey outlines the research in activity recognition with main focus on the deep model used for processing the data from sensors.

\subsection{Wearable Device Based}
 This section presents the surveys which discuss wearable device based solutions for activity recognition.

Lara \& Labarador \cite{RN80} outlined the work conducted in activity recognition using wearable sensors. This survey presents a detailed discussion of different design issues in a HAR system, such as selecting sensors and attributes, data collection and protocol, recognition performance, processing methods, and energy consumption. This survey categorizes the existing work into supervised online, supervised off-line, and semi-supervised off-line systems. Human activity recognition systems which use wearable sensors for data collection is the main focus of this survey.

Cornacchia et al. \cite{RN89} presented a detailed survey and divides the existing research work in two major categories: global body motion activity, which involves the movement/displacement of the whole body (e.g., walking, climbing, and running) and local interaction activity, which involves the movement of extremities (e.g., use of objects). This paper also provides a classification based on the type of sensor used and the placement of the sensor on the human body such as a waist mounted and a chest mounted. Different techniques using sensors like gyroscope, accelerometer, magnetometer, wearable cameras, and hybrid sensors (using multiple sensors) have been discussed by the authors. This survey focuses only on wearable sensors based research work for activity recognition.

Some surveys also focus on mobile phone-based solution for HAR because many techniques use a mobile phone (built-in sensors) for activity recognition. One such survey is presented by Shoaib et al. \cite{RN85} which outlines the research work using mobile phones.

\begin{normalsize}
\begin{center}
\begin{table*}[!t]
\centering
\caption{\label{tab: 4} Summary of the previous surveys}
\begin{adjustbox}{width=\textwidth}
\begin{tabular}{|p{22mm}|p{27mm}|p{5cm}|p{2cm}|p{25mm}|}
\hline

\textbf{Categories}                    & \textbf{Paper}    & \textbf{Main Focus}                                    & \textbf{Future Research Directions} & \textbf{Comparisons of Different Techniques} \\ \hline
\multirow{5}{*}{\begin{minipage}{2.2 cm}RF Based\end{minipage}}              & Scholz et al. \cite{RN78}      & Applicability of radio sensors in activity recognition & Yes                                 & No                                           \\ \cline{2-5} 
                                       & Amendola et al. \cite{RN81}    & Applications of RFID technology in various fields      & Yes                                 & No                                           \\ \cline{2-5} 
                                       & Wang and Zhou \cite{RN83}      & Use of radio signals for activity recognition          & Yes                                 & Yes                                          \\ \cline{2-5} 
                                       & Ma et al. \cite{RN86}          & Wi-Fi based techniques                                 & No                                  & No                                           \\ \cline{2-5} 
                                       & Cianca et al. \cite{RN87}     & FM radio and Wi-Fi based methods                       & No                                  & No                                           \\ \hline
\multirow{2}{*}{\begin{minipage}{3 cm}Sensor Based\end{minipage}}          & Chen et al. \cite{RN79}        & Data-centric activity recognition techniques.          & Yes                                 & Yes                                          \\ \cline{2-5} 
                                       & Wang et al. \cite{RN90}        & Deep models for sensor based approaches                & Yes                                 & Yes                                          \\ \hline
\multirow{3}{*}{\begin{minipage}{2.2 cm}Wearable Device Based\end{minipage}} & Lara \& Labarador \cite{RN80}  & Wearable sensors based approaches                      & Yes                                 & Yes                                          \\ \cline{2-5} 
                                       & Shoaib et al. \cite{RN85}      & Mobile Phones based techniques                         & Yes                                 & Yes                                          \\ \cline{2-5} 
                                       & Cornacchia et al. \cite{RN89}  & Wearable sensor based techniques                       & No                                  & Yes                                          \\ \hline
\multirow{2}{*}{\begin{minipage}{3 cm}Vision Based\end{minipage}}          & Vrigkas et al. \cite{RN84}     & Vision based approaches                                & Yes                                 & Yes                                          \\ \cline{2-5} 
                                       & Herath et al. \cite{RN88}      & Vision based solutions                                 & Yes                                 & Yes                                          \\ \hline

\end{tabular}
\end{adjustbox}
\end{table*}
\end{center} 
\end{normalsize}

\vspace{-3.4mm}
\subsection{Vision Based}
This section presents the surveys which focus on vision-based solutions for activity recognition.
Vrigkas et al. \cite{RN84} presented a survey of existing research work which uses vision-based approach for activity recognition and classified the literature in two main categories: unimodal and multi-modal approaches. Unimodal methods are those which use data from a single modality and are further classified as stochastic, rule-based, space-time based, and shape-based methods. Multi-modal approaches use data from different sources and are further divided into behavioral, effective, and social-networking methods. This survey focuses only on vision-based approaches for activity recognition.

Herath et al. \cite{RN88} provided a detailed overview of the major research undertaken in the field of action recognition, using vision-based approaches. This survey categorizes the overall work into two major categories: solutions based on representation and solutions based on deep neural network. Representation-based solutions are further classified into Holistic and local presentations and aggregation methods. Solutions based on the deep neural network are sub-classified as multiple stream networks, temporal coherency networks, generative models, and spatiotemporal networks. This paper provides a very detailed and comprehensive survey of the work done in action recognition.

Surveys discussed above are summarized in Table 1. Most of these surveys highlight the work conducted in human activity recognition but the focus is mainly on a single approach. Some focus on sensor-based approach while others focus on vision-based approach. Also, these surveys do not provide details about the weaknesses and strengths of different approaches for activity recognition.

Recently, with the advancement in RFID technology, many solutions have been proposed for activity recognition using device-free RFID technology. Previous surveys missed the details about these solutions. To the best of our knowledge, there is no previous survey which provides a comprehensive and detailed analysis of RFID-based device-free approaches for activity recognition. Our goal is not only to provide an overview of the latest research conducted in human activity recognition with main focus on device-free approaches, especially RFID, but also to compare different techniques and understand the advantages and disadvantages of each technique.

\section{Technical Background}

Human activity recognition is a composite process and can be divided into four major phases  \cite{RN79} as shown in Figure 2. These phases are: i) selection and deployment of sensors ii) collection of data from these sensors, iii) pre-processing and feature selection from the data, and iv) use of machine learning algorithm to infer or recognize activities.

Over the past decade, considerable research has been done in the area of human activity recognition using sensor technology. Some of the most common sensors which are used for activity recognition are: accelerometer, motion sensors, biosensors, gyroscope, pressure sensor, proximity sensor, etc.  Some of the sensors are radio-based such as RFID. These sensors can be used in various ways. They can be attached to different objects or can be used as wearable sensors or be deployed in the environment.

Nowadays, many different kinds of cheap and portable sensors are available which have the ability to sense and communicate the information using wireless networks. In this section, we provide the details about some of the technologies which are being used for human activity recognition as shown in Figure 3 and 4.

\subsection{Surveillance Cameras}
The most basic and traditional way of activity recognition is to install surveillance cameras in the facility and monitor the activities of humans. Monitoring can be done through human (a person watching the videos and images coming from cameras) or through an automatic process. Different computer vision techniques have been developed which can process and analyze the data (videos and images) from the camera and can automatically recognize activities. 

\begin{figure}[hbt!]
	\centering
	\includegraphics[width= \linewidth]{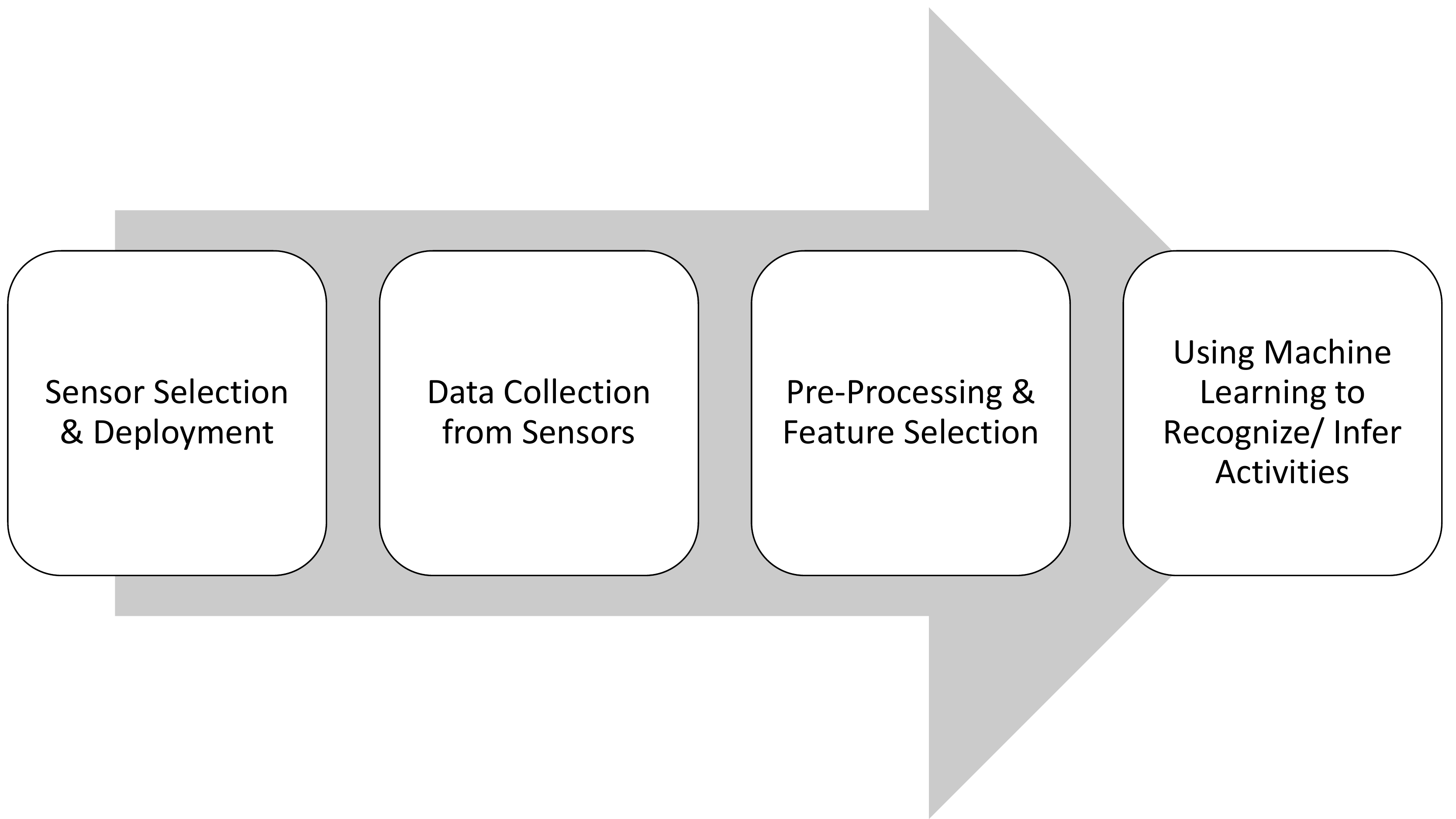}
	\caption{General process of human activity recognition}
	\label{fig2}
\end{figure}

\begin{figure}[hbt!]
	\centering
	\includegraphics[width=  \linewidth]{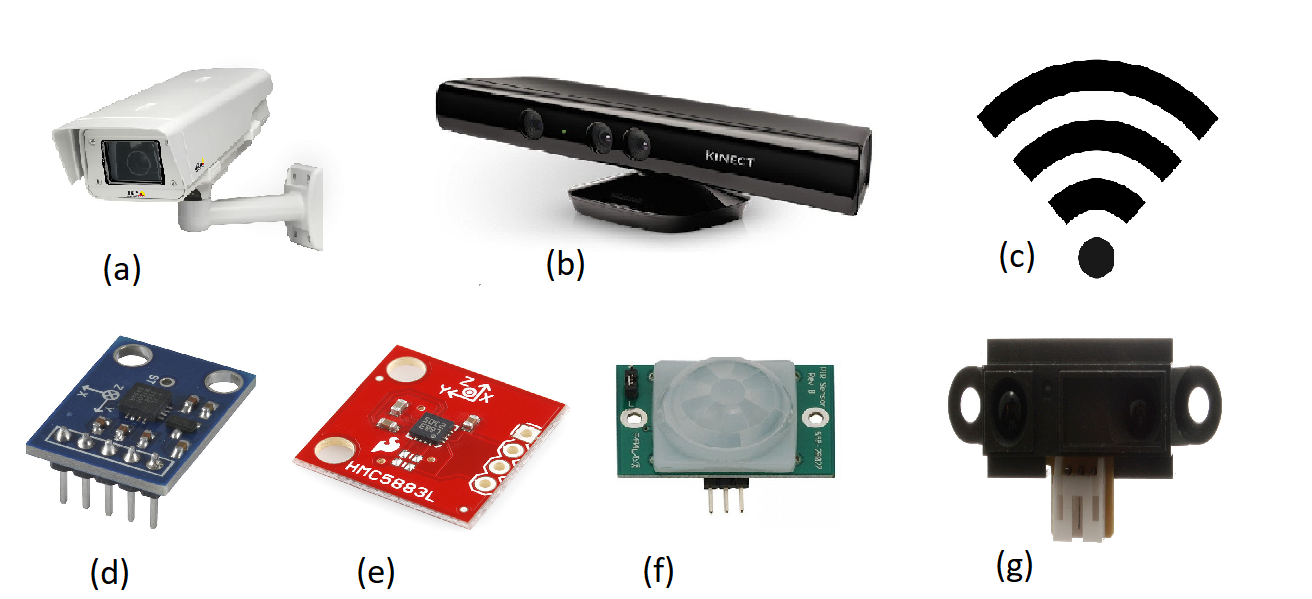}
	\caption{Some technologies used for activity recognition: (a) Surveillance camera (b) Depth camera (c) Wi-Fi (d) Accelerometer                      (e) Magnetometer (f) Motion sensor (g) Proximity sensor}
	\label{fig1}
\end{figure}

\begin{figure}[hbt!]
	\centering
	\includegraphics[width=  \linewidth]{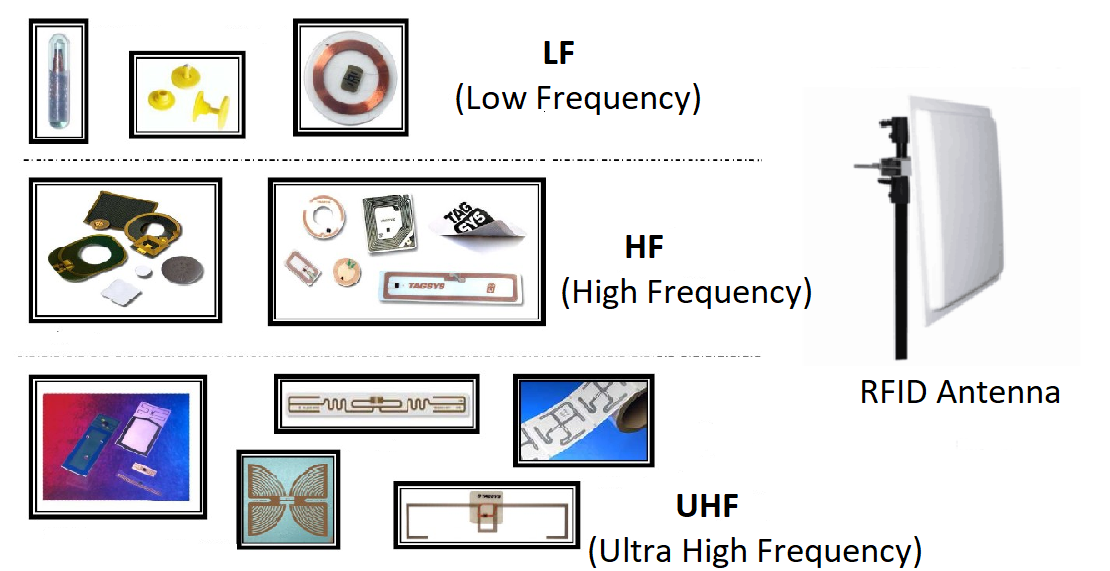}
	\caption{RFID technology; tags and antenna}
	\label{fig1}
\end{figure}

\subsection{Depth Cameras}
One of the issues with traditional cameras is dependency on light i.e., they cannot work in darkness. The development of depth cameras such as Kinect solved this issue because it can work in total darkness. Different data streams can be obtained from Kinect such as RGB, depth, and audio \cite{maret2018real}. It can capture the information about human body and can construct a 3D virtual skeleton. Using this information, activities can be recognized because different movements of the body (skeleton) are related to different activities. Apart from complex computation, cost of depth cameras is high, which is a disadvantage of using it for activity recognition.

\subsection{Wi-Fi}
In the last decade, there is a paradigm shift in human activity recognition research from device-bound approaches to device-free approaches. Researchers have studied the properties of wireless networks, such as Channel State Information (CSI) and started to use it for activity recognition \cite{RN86}. Many solutions have been proposed for localization, tracking, fall detection, etc., using Wi-Fi. A major advantage of Wi-Fi is that it is unobtrusive and users are not required to carry any device with them.

\subsection{Sensors}
In the twenty-first century, significant research has been done in the field of sensors and many different kinds of sensors have been produced. These sensors are very useful and have the ability to sense the environment and communicate the information wirelessly. Some of the sensors which are widely used in the research for activity recognition are given as follows.

\subsubsection{Accelerometer}
An accelerometer is an electromechanical device used for measuring the acceleration. It can sense acceleration in multiple directions. To do that, the accelerometer is designed with multi-axis (i.e., x, y, and z) sensors. A multi-axis accelerometer can measure acceleration in x, y, and z-direction at the same time. The accelerometer is widely used in solutions for gesture recognition, posture recognition, fall detection, tracking, ambient assisted living, activities of daily living, etc.

\subsubsection{Magnetometer}
Magnetometer is used to measure the magnetic field and some time the direction of magnetic field. This sensor is used in various fields of activity recognition (e.g., gesture recognition) because of its ability to detect changes in the magnetic field caused by human activity.

\subsubsection{Motion Sensor}
Motion sensors are used to detect the motion or presence of a subject in a particular area. Motion sensor are widely used in the field of human activity recognition especially in motion detection, tracking, and people counting. 

\subsubsection{Proximity Sensor}
It is an electronic sensor which can detect the presence of nearby objects without making any physical contact. Proximity sensors are widely used in gesture recognition techniques.

\subsection{RFID}
Radio Frequency Identification technology has seen a boom in the last decade. Originally developed for military purposes to differentiate between friendly and hostile aircrafts \cite{landt2005history}, this technology has seen momentous advancement in recent years \cite{wu2011rfid}. It is widely used in tracking and supply chain management. Initially, the range of RFID technology was very small (few centimeters) which is now increased up to a great extent (15 meters for passive tags and 100 meters for active tags) \cite{ko2017accessibility}. The RFID technology has two main parts; reader and tags.   
\vspace*{-1mm}
\begin{itemize}
	\itemsep -1pt
	\item 
Reader is a device which is used to collect information from tags. The reader has an antenna which emits radio waves. These radio waves are received and modulated by RFID tags with their information such as ID. The reader can capture these backscattered signals through an antenna, which has the information of tags. 
	\item 
	Tags are the small chips with which can be easily attached to any objects. These tags are mainly of two types; Active and Passive. Active tags have their own power supply (battery) while passive tags are battery less and harvest their energy from the radio waves of the readers. Active tags have longer range as compared to the passive tags.	
\end{itemize}
Due to the passive nature of RFID, low cost and unobtrusive, this technology have been adopted in various fields. RFID is now widely used in human activity recognition research. Researchers are using RFID technology for posture recognition, gesture recognition, tracking, localization, behavior recognition, etc.

\begin{figure*}[]
	
	\includegraphics[width= \linewidth]{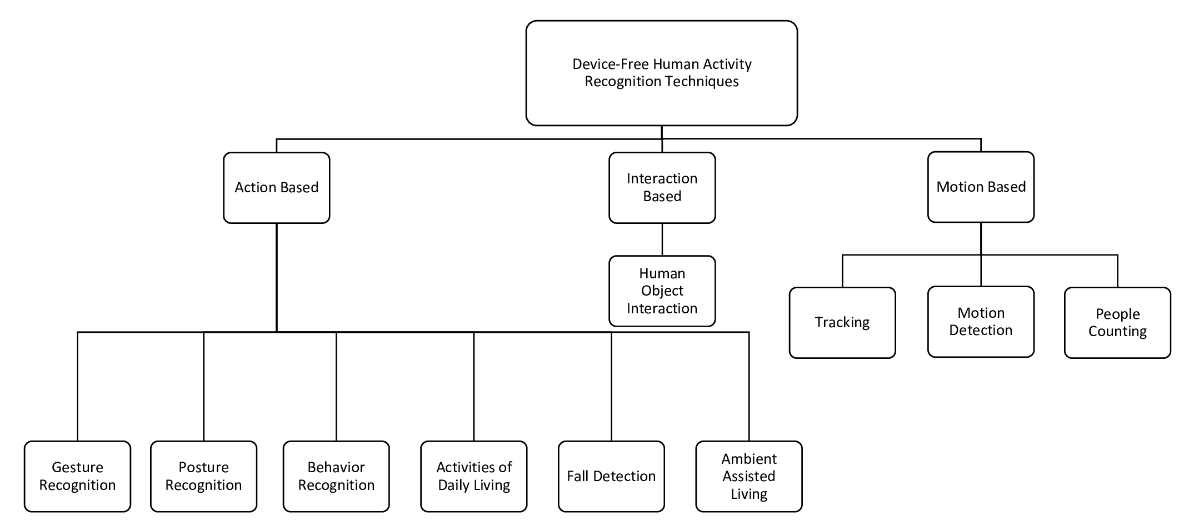}
	\caption{Overview of human activity recognition techniques which can be divided into three main categories and then into 10 sub-categories.}
	\label{fig1}
\end{figure*}

\section{Device-Free Human Activity Recognition Techniques}

	Activity recognition aims to identify or detect physical activities of a single person or group of persons. These physical activities can be of different types. Some of these activities can be performed by a single person which involves the movement of the whole body such as walking, running and sitting. Some of these activities can be complex like jumping and dancing. Some activities involve a specific body part such as making gestures with hand. Certain activities can be performed by interacting with objects, for example, preparing a meal in the kitchen. Detecting the presence or motion of a human in a certain environment also comes under the activity recognition (e.g., intrusion detection). Tracking the movement or trajectory of a human in a specific area can also be considered as activity recognition. Significant research has been conducted under the umbrella of human activity recognition. A schematic classification of different works in the field of activity recognition is given in Figure 5. In this paper, we will follow the taxonomy given in Figure 5 and will provide an overview of the research work done in these areas with a focus on device-free techniques, especially RFID technology. 

\subsection{Comparison Metrics}
\label{sec:cmp_metrics}
Prior to the discussion of different categories of device-free human activity recognition techniques, we provide the comparison metrics in this section and these metrics will be used in Section \ref{sec:action-based}, Section \ref{sec:motion-based} and Section \ref{interaction-based} for the comparison of different approaches. Following are the metrics that we have used for comparing various solutions presented in this survey.

\begin{itemize}
	\itemsep -1pt
	\item 
	 \textbf{Approach (M1)}: 
     Various approaches have been used by researchers for human activity recognition. These approaches can be device-free, wearable or hybrid. Hybrid approaches combine both wearable and object-tagged approaches. We have listed the approach used by solutions presented. \textit{D} represents device-free approach, \textit{W} represents wearable approach, whereas \textit{H} represents hybrid approach.  
	\item 
	\textbf{Technology (M2)}:
    Literature shows that different solutions have used different technologies. Some of the prominent technologies used in the area of HAR are RFID, Kinect, Infra-Red, Radar, Sensor Fusion, Wi-Fi, Hybrid (fusion of multiple technologies), etc. We have listed the technology used by different solutions.   
    \item 
    \textbf{Information Type (M3)}:
    Different techniques use different information as input for performing the required task. Solutions using the same approach and technology can use different information as input. We have provided the type of information used by different techniques as input for their processing.  
    \item 
    \textbf{Machine Learning Algorithm Used (M4)}:
    Machine learning is an essential part of the process of human activity recognition. Different types of machine learning algorithms have been used in HAR. Some of the most famous algorithms are Support Vector Machine (SVM), \textit{k}-Nearest Neighbour (KNN), Random Forest (RF), Hidden Markov Model (HMM), Naive Bayes (NB), Decision Tree (DT), etc. We have given the machine learning tool used by the techniques presented.   
    \item 
   \textbf{ Supervised/Unsupervised (M5)}:
   Machine learning algorithms can be supervised or unsupervised. Both are different approaches. Supervised techniques need training data while unsupervised techniques do not need any training data. We have provided this information for the presented papers.\textit{Y} represents supervised whereas \textit{N} represents unsupervised.
    \item 
    \textbf{Application (M6)}:
    Human activity recognition is a very vast field. Different techniques focus on different applications. Some provide the solution for gesture recognition while others provide solution for tracking. We have provided the applications areas for the presented papers. 
    \item 
   \textbf{ Cost (M7)}
   Cost is a key factor for any technique. If accuracy of a  solution is good but cost is too high, then it's of no practical use. We have provided information about the cost of the techniques discussed. Cost is divided into two categories: expensive (if device per person is used) and cheap (if single device is used for all participants).  
    \item 
    \textbf{Accuracy (M8)}
    A very important factor for the evaluation of a solution is its accuracy. We have provided information about the accuracy of the given techniques. We have categorized the accuracy in three categories; High ( $\textgreater$ 90\% ), Medium ( $\textgreater$ 80\% \& $\textless$ 90\% ) and Low ( $\textless$ 70\% )
    \item 
    \textbf{Latency (M9)}
    Latency is a critical factor, especially for real time applications. If a solution is accurate but takes long time to provide the results, it is not practical. We have provided the latency information about the presented solutions.
    \item 
    \textbf{Real-time (M10)}
    Last but not least is whether a solution is real-time or not. This is important for human activity recognition because getting the results in real time is a compulsion in many situations. For example, in the case of gesture recognition, it is required to get the results in real time. We have included this factor in our comparison table. \textit{Y} means the solution is real time whereas \textit{N} means the solution is not real time.       
\end{itemize}

\subsection{Action Based Activities}
\label{sec:action-based}
Action based activities are those activities which involve some action of the human body. This action can involve either the whole body or a specific portion of a body. In this section, we provide an overview of the different solutions proposed for the recognition of action based human activities.

\begin{figure}[h]
	\centering
	\includegraphics[width=  3 in]{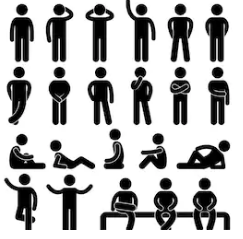}
	\caption{Examples of action based activities.\protect\footnotemark}
	\label{fig1}
\end{figure}

\footnotetext{Source: https://www.shutterstock.com/image-vector/man-basic-posture-people-sitting-standing-88609435}

\begin{normalsize}
\begin{table*}[!t]
\centering
\caption{\label{tab: 4} Approach, technology, advantages, disadvantages and applications of the techniques presented for gesture recognition }
\begin{adjustbox}{width=\textwidth}
\begin{tabular}{|p{2cm}|p{3cm}|p{2cm}|p{3.5cm}|p{3.5cm}|}
\hline
\textbf{Approach} & \textbf{Technology}                          & \textbf{Advantages} & \textbf{Disadvantages}                       & \textbf{Applications}                                                                                             \\ \hline
Vision Based      & Surveillance camera                          & High accuracy        & High cost, complex computation, privacy issue & \multirow{6}{*}{\begin{minipage}{3.5 cm}Gaming,\\ Smart screen interaction,\\ Sign language interpretation,\\ Remote monitoring \end{minipage}} \\ \cline{1-4}  
Depth sensor      & Kinect                                       & High accuracy        & High cost, privacy issue                      &                                                                                                                   \\ \cline{1-4}
Wearable Sensors  & Gloves, Bracelet, Smart Watch                & Low cost             & Constraint to wear the device                 &                                                                                                                   \\ \cline{1-4}
Object Tagged     & Accerelometer, Ultrosonic Sensor, Microphone & Low cost             & Device-bound                                  &                                                                                                                   \\ \cline{1-4}
RFID              & Passive RFID tags arrays                    & Low cost, passive    & Environmental interference                    &                                                                                                                   \\ \cline{1-4}
Radio Frequency   & Radar, Wi-Fi                                 & Low cost             & Environmental interference                    &                                                                                                                   \\ \hline

\end{tabular}
\end{adjustbox}
\end{table*}
\end{normalsize}  

\subsubsection{Gesture Recognition}
Gesture recognition is one of the most important sub-topics in action recognition. In recent years, it has gained much attention for its role in human-machine interaction. In the past, the only option to interact with a machine was through manually using some device such as a mouse,  keyboard or touch screen. But that is not always feasible. For example, using these devices may not be possible for the disable or elderly people. Placing these input devices in public spaces such as parks, airports, and hospitals may not be easy because of the associated high cost or heavy usage. Also, due to their touch-based nature, it can be harmful to use them in some environments such as hospitals, where the chances of infections are high.

Over the past decade, researchers are trying to provide alternative solutions for interacting with machines. Table 2 provides the approach, technology, advantages, and disadvantages of the different techniques discussed in this section and some applications of gesture recognition. Some of these solutions are vision-based and use cameras to capture videos or images for gesture recognition \cite{RN57,RN58,RN59,RN60, RN61,RN62,RN63,RN64}. But this approach has a privacy issue, complex processing, and very high deployment cost. Some solutions use wearable devices for gesture recognition. These devices range from simple sensors to specially designed gloves and bracelets \cite{RN65,RN67,RN68, RN66}. Some techniques use objects tagged with sensors and users make gestures with these sensor-tagged objects, which can be recognized \cite{RN69,RN70,RN71}. Jayatilaka \& Ranasinghe \cite{RN72} used a smart cup tagged with passive RFID tags to recognize fluid intake gestures while another work tried to identify drinking gestures by using a wrist-worn sensor \cite{RN73}. The focus of this section is device-free RFID-based solutions for gesture recognition.

Ye et al. \cite{RN74} proposed a device-free solution called Link State Indicator (LSI), for gesture recognition, using passive RFID tag arrays. It uses the number of counts (tag being read successfully), to represent the state of an unobstructed link. LSI is the ratio of the tag's count read successfully to a reference count in a unit time. The reference count is obtained when there is no obstruction.  For each gesture, this work calculates gesture matrix which represents the state for all the tags as fully obstructed, partially obstructed or not obstructed at all. Finally, Fisher's linear discriminant method is used for gesture recognition. To evaluate the performance of this scheme, the authors have performed experiments using twelve different gestures: six gestures while squatting and six gestures while standing. The system has an average accuracy of 94\% for all the twelve gestures. This technique is off-line, so it cannot provide real-time recognition. The gestures identified are very different from each other and the performance is poor for closely related gestures. The paper lacks the discussion about the complexity (space and time) of the proposed solution. Also, the number of experiments performed is not enough. There is no discussion about the issue of variability i.e. if the same gesture is performed by different persons or the same person performs the same gesture in different styles.

Smart surface \cite{RN75} is a technique which combines RFID technology with machine learning for recognizing gestures. This technique uses passive RFID tags and antennas attached to a surface. The basic idea of this work is that the Received Signal Strength Indicator (RSSI) values from RFID tags are disturbed when a gesture is performed in front of the tags. These disturbances can be classified by the K-means algorithm into different clusters with each cluster representing a specific gesture performed. These specific gestures can be used with many different applications such as controlling an audio player. Experiments show that smart surface can achieve 100\% accuracy for some basic gestures (e.g., hand movement from 1 tag to another) when separate antennas are used with each tag. This work provides an initiative for making smart surfaces using passive RFID tags but it can identify very basic gestures only, which are some movements made in pre-defined directions. Every tag is required to have its own antenna (e.g., 10 tags will need 10 antennas). Also, this work lacks the details about the use of the system such as how far should be the user from the tags, while performing gestures.

Ding et al. \cite{RN76} proposed a device free technique for gesture (hand motion and handwriting) recognition using passive RFID tags. This technique uses COTS RFID tags attached to a plate in a grid form. The system is based on the idea that when a motion (hand gesture) occurs in front of an RFID tag, significant change can be seen in the RSSI and phase values received by the reader. Using these changes in RSSI and phase values combined with tag IDs, different hand gestures can be identified. The system is capable of identifying some basic gestures for touchscreen as well as English alphabets i.e., English letters drawn with a hand motion in the air can be recognized. Experiments show that the proposed system achieves an accuracy of 91\%. One of the best aspects of this technique is that no prior training is required and it can provide results in real time which is important for interactive applications. The system works well when the gestures are performed at relatively slow speed but performance degrades when gestures are performed with high speed. Another limitation of this technique is the distance from the surface (plate with tags). User needs to be very close ($\leq$ 5 cm) to the plate while making any gestures otherwise, performance degrades.

GRfid \cite{RN77} is a device-free approach for gesture recognition. It is capable of detecting a total of six hand gestures. The system uses the RFID signal phase changes for recognizing different hand gestures. Data collected from passive RFID tags is passed through several processing blocks namely, pre-processing, gesture detection, gesture profiling training, and gesture recognition. GRfid uses Dynamic Time Warping (DTW) as a metric for comparison and proposes an adaptive weighting algorithm for gesture matching. The achieved accuracy is 96.5\% for the identical position (testing and training data is collected at the same position) and 92.8\% for diverse-position (testing and training data is collected at different positions) scenarios. Although the proposed system performed better in many diverse scenarios, there are some limitations. The gestures tested and recognized are very basic in nature. There is no discussion about the latency of the system which is a very important aspect for the gesture recognition system.

\begin{normalsize}
\begin{table*}[h!]
\centering
\caption{\label{tab: 4} Approach, technology, advantages, disadvantages and applications of the techniques presented for posture recognition}
\begin{adjustbox}{width=\textwidth}
\begin{tabular}{|p{2.5cm}|p{3cm}|p{2.5cm}|p{3.5cm}|p{2cm}|}
\hline

\textbf{Approach}            & \textbf{Technology}                  & \textbf{Advantages}              & \textbf{Disadvantages}                       & \textbf{Applications}                                                \\ \hline
Vision Based                 & Camera                               & High accuracy                     & High cost, complex computation, privacy issue & \multirow{5}{*}{\begin{minipage}{2 cm}Smart homes,\\ Smart offices,\\ Hospitals,\\ Care centers\end{minipage}} \\ \cline{1-4}
Wearable devices             & Smartphone, accelerometer, gyroscope & Low cost                          & Constraint to carry the device                &                                                                      \\ \cline{1-4}
\multirow{3}{*}{Device Free} & RFID                                 & Low cost, COTS available, passive & Environmental interference                    &                                                                      \\ \cline{2-4}
                             & Radar                                & Low cost                          & Customized hardware required                  &                                                                      \\ \cline{2-4}
                             & Wi-Fi                                & Low cost, COTS available          & Environmental interference                    &                                                                      \\ \hline

\end{tabular}
\end{adjustbox}
\end{table*}
\end{normalsize}

\subsubsection{Posture Recognition}
Humans do many activities in their daily lives. These activities can be simple postures such as standing, sitting, lying or walking or may be complex such as running, doing exercise, and cooking. Many of these simple activities (postures) are of interest to recognize because of many applications in various fields. Table 3 provides the approach, technology, advantages and disadvantages of the different techniques discussed in this section and some applications of posture recognition. Researchers have used different sensor-based techniques for posture recognition and these techniques can be broadly classified in two main categories: i)  using on-body or wearable sensors, and ii) using sensors deployed in environment (device free). First category of solutions use wearable sensors i.e., different sensors are attached to human body or clothes while performing activities. Some solutions use inertial sensors embedded in smartphone i.e., users need to carry their smartphone while performing any activity. A smartphone based technique is presented in \cite{torres2015accelerometer} which uses accelerometer sensor embedded in smartphones. This technique is fully implemented in the android smartphone and allows for different orientation and placement of smartphone on human body. Wickramasinghe \& Ranasunghe \cite{RN22} presented a technique using wearable sensor for ambulatory monitoring to recognize activities such as transfer out of chair, bed or walking. This method requires the user to wear a computational Radio Frequency Identification (CRFID) sensor. Ronao \& Cho \cite{RN23} proposed a deep neural network based solution using embedded sensors in smartphone (accelerometer and gyroscope).  Many other solutions have been presented for posture recognition using different wearable sensors \cite{RN25,RN26,RN27,RN28}. A major problem with wearable sensors is that it is not always feasible to wear these tags, while performing an activity.

A more realistic approach is the device-free approach. In the past decade, many solutions have been proposed for posture recognition using the device-free approach. RF-Care \cite{RN13} proposed a device-free solution for posture recognition based on RFID technology. The passive RFID tags arrays are placed in the environment to capture the activity information. When a posture is performed in front of these tag arrays, the disturbance causes variation in the RSSI values of these tags. RF-Care analyses and uses these changes for posture recognition. This work also studies the issue of tags placement in an indoor environment and provides an optimal setting for the tag array's deployment to achieve the best results with minimum computation cost. Various selection techniques are proposed and compared such as relief-F, F-statistic and random forest. RF-Care uses Support Vector Machine (SVM) for recognition of steady postures and for posture transition detection, Hidden Markav Model (HMM) is used. The proposed system is deployed in an indoor environment and different experiments are conducted in two scenarios (office and home) to evaluate the performance of RF-Care. In the presence of one subject, the system achieves an accuracy of 98\% in steady posture recognition for both the scenarios, using 9 tags and one reader. For posture transition, the proposed system achieves an accuracy of 70\%. RF-Care provides a very simple and easy to implement solution but it has a latency of around 3.5 seconds which may be too long for some applications such as interactive environments. The accuracy for posture transition detection needs to improve. The proposed solution needs to be evaluated to check the effect of interference from the environment such as obstacle lying in the area of the presence of other people.

Yao et al. \cite{RN19}  presented a device-free RFID-based technique for activity recognition. This work combines machine learning with RFID technology and proposes a dictionary-based approach which can learn the dictionaries for different activities in an unsupervised manner. The system uses RFID tags deployed in arrays for capturing activity information. Raw data from the tags is passed through a segmentation process in which the continuous sequence is divided into individual segments. Each segment represents a specific activity. The paper proposes and uses a sliding window segmentation algorithm which is based on slope variation. Seven features are selected by using a ranking method based on canonical correlation analysis. For activity recognition, this technique uses a sparse dictionary-based approach, in which a single dictionary is learned for each activity. The authors have deployed the proposed system in real environments and the comparisons with other approaches show that the proposed system can achieve better accuracy. A limitation of this work is the latency i.e., it takes around 4.5 seconds for recognition of an activity which may be too slow for some applications.

\begin{normalsize}
\begin{table*}[h!]
\centering
\normalsize
\caption{\label{tab: 4} Approach, technology, advantages, disadvantages and applications of the techniques presented for behaviour recognition}
\begin{adjustbox}{width=\textwidth}
\begin{tabular}{|p{2cm}|p{2cm}|p{4cm}|p{6cm}|p{3.5cm}|}
\hline

\textbf{Technologies}         & \textbf{Examples}   & \textbf{Advantages}              & \textbf{Disadvantages}                       & \textbf{Applications}                                                                   \\ \hline
Vision         & Surveillance camera & High accuracy                     & High cost, complex computation, privacy issue & \multirow{4}{*}{\begin{minipage}{3.5cm}Shopping centers,\\ Theme parks,\\ Care centers,\\ Security \& Surveillance\end{minipage}} \\ \cline{1-4}
Depth sensor                 & Kinect              & High accuracy                     & High cost, privacy issue                      &                                                                                         \\ \cline{1-4}
\multirow{2}{*}{Device free} & Wi-Fi               & Low cost, COTS available          & Environmental interference                    &                                                                                         \\ \cline{2-4}
                             & RFID                & Low cost, COTS available, Passive & Environmental interference                    &                                                                                         \\ \hline

\end{tabular}
\end{adjustbox}
\end{table*}
\end{normalsize}

Yao et al. \cite{li2018r} presented a solution called R\&P, which is device-free and uses passive RFID technology for human activity recognition. R\&P extracts phase and RSSI values from the RFID tags deployed in the environment and uses these values for activity recognition. Unlike some other RFID-based techniques, this work combines both RSSI and phase values for recognition. For de-noising of RSSI and phase values, D-Gaussian algorithm \cite{RN17} and stein unbiased risk estimate based method \cite{RN18} are used, respectively. R\&P uses the DTW algorithm for feature matching and proposes a modified version of DTW called T-DTW, which can reduce the matching time by 60\%. To evaluate the performance of R\&P, various experiments with different settings, are done by the authors. The proposed technique shows good results in different realistic scenarios such as empty hall, office and book bar. The system is capable of identifying six activities with an average accuracy of 87.9\%. It is a simple approach which can achieve good results but there are some limitations. It lacks the discussion about the complexity and time requirement for activity recognition. Tested activities are very different from each other and the system needs to be evaluated for activities which are very similar to each other such as standing \& walking or sitting \& lying.

Recently, an RF-radar based approach was presented by Avrahami et al. \cite{avrahami2018below} for recognizing human activities in a checkout counter of a convenience store and a typical office desk.  The proposed technique uses Walabot Pro sensor which is an RF-radar with 18 antennas and is capable of constructing a 3D image from the reflected radio waves. This sensor is deployed under the work surface and when the subject performs pre-defined activities, data is captured in the form of RF samples. For comparison purpose, the proposed system also uses a wearable IMU sensor (Microsoft Band 2) and data is captured from the IMU sensor during the experiments. Different techniques such as SVM, Random Forest, and Naive Base, are used for classification of performed activities. Experiments in both scenarios (checkout counter and office desk) prove that RF-radar can perform better than IMU and by combining both approaches, accuracy can further be improved. By increasing the projections of RF-radar, the accuracy of recognition can be increased. The proposed system achieves an accuracy of 95.3\% for office desk-work and 34.9\% for the checkout counter.

\subsubsection{Behavior Recognition}
Behavior recognition is an important sub-area of human activity recognition. The basic idea is to infer/recognize the behavior of a person from the data captured through different sensors. Behavior recognition is very useful in various scenarios such as smart environments (elderly care centers and smart homes) \cite{RN91} and shopping centers. In elderly care centers, patients can be monitored remotely which can reduce the cost significantly because human resources are very expensive.  Any abnormality in the behavior of elder people can be detected and the concerned people can be informed of the situation. In shopping centers, behavior identification of the customers can help owners to improve their business. Customer's shopping information such as interests, preferences, and brands, can be very useful to further improve the shopping experience for the customers. 
Recently, considerable work has been done to identify the behavior of customers while shopping. Some applications of behavior recognition along with the approach, technology, advantages, and disadvantages of various techniques presented in this section are given in Table 4. One study proposes a technique, based on the surveillance system, for analysis of shopping behavior \cite{RN92}. The given system uses multiple cameras to track the movements of customers. One other technique used a Kinect sensor for behavior recognition of customers \cite{RN93}. Besides problems such as computation complexity and cost, privacy is a major issue with vision-based approaches. Zeng et al. \cite{RN94} proposed a Wi-Fi-based technique using CSI to recognize the behavior of customers while they shop. The given system is capable of detecting coarse-grained activities only, such as standing, walking and walking fast. The reason is, CSI cannot provide enough information to recognize fine-grained activities e.g., the customer is just looking at a specific item, the customer is looking in detail and is interested or customer is putting the item in cart.

Nowadays, researchers are using passive RFID technology for recognition of shopping behavior. Han et al.  \cite{RN96} proposed a behavior identification system called Customer Behavior Identification (CBID). The given system can analyze the wireless signal collected from RFID tags, attached to different items in the shopping center. CBID is capable of detecting popular item (item picked by most customers), an explicit correlation between items (rivalry or complementary), and implicit correlation between items (items picked or purchased at the same time). CBID uses phase changes and Doppler frequency shift, which occurs as a result of the movement of the items. The authors implemented a prototype of the proposed system and performed extensive experiments to evaluate the performance of CBID. CBID achieves good results in all realistic scenarios but the proposed system needs to be evaluated for metallic products to check the effect of interference with the signal. Also, this work lacks the discussion about the system latency.

\begin{normalsize}
\begin{table*}[h!]
\centering
\caption{\label{tab: 4} Approach, technology, advantages, disadvantages and applications of the techniques presented for fall detection}
\begin{adjustbox}{width=\textwidth}
\begin{tabular}{|p{2cm}|p{3.8cm}|p{2.7cm}|p{2.5cm}|p{2.5cm}|}
\hline

\textbf{Approach}            & \textbf{Technology}                                       & \textbf{Advantages}               & \textbf{Disadvantages}        & \textbf{Applications}                                                \\ \hline
Wearable device              & Accelerometer + RFID, smartphone, barometer, magnetometer & Low cost                          & Constraint to wear the device & \multirow{4}{*}{\begin{minipage}{2.5 cm}Elder care centers, \\Hospitals, \\Industrial workplace\end{minipage}} \\ \cline{1-4}
\multirow{3}{*}{Device free} & Wi-Fi                                                     & Low cost, COTS available          & Environmental interference    &                                                                      \\ \cline{2-4}
                             & Radio devices                                             & Low cost                          & Customized hardware required  &                                                                      \\ \cline{2-4}
                             & RFID                                                      & Low cost, COTS available, passive & Environmental interference    &                                                                      \\ \hline

\end{tabular}
\end{adjustbox}
\end{table*}
\end{normalsize}

Zhou et al. \cite{RN97} tried to solve the problem of customer shopping behavior mining by using COTS passive RFID tags. Passive RFID tags are attached to different items of the store. When users interact with these items, significant changes occur in phase readings of these tags. The proposed system exploits these changes for mining the customer's behavior. The system is capable of detecting different actions of a customer such as browsing through items, picking an item and trying items together (correlation between items). The basic idea is, when a user is just passing by a rack (browsing), phase values will be disturbed slightly and when an item is picked by a user (showing interest), phase readings will change significantly. When multiple items are tried together, these can be detected by finding the correlation between the tags. The authors implemented a prototype of the proposed system in a realistic scenario to evaluate the performance. It achieves good results for detecting popular category (items most browsed), hot items (items picked by customers) and correlated items (items picked together by customers). Performance of the given system degrades in a crowded store, where a large number of customers are shopping.

 \subsubsection{Fall Detection}
Fall means when the position of the human body suddenly changes from the normal state (e.g., standing, sitting or walking) to reclining, without any control \cite{RN33}. Fall can result in injuries both minor and major. Around 3\% of falls result in fractures but even minor injuries are not good as these may cause delays in the rehabilitation of the patient and can also cause further stress \cite{RN35}. Elder people are more vulnerable to falls \cite{RN36} and can face severe injuries even death as a result of fall \cite{RN34}. Besides injuries, falls can add a lot to medical expenses. For the year 2015, in the USA alone, the billing cost for falls was around \$50 billion \cite{RN36} and as reported by Center for Disease Control and Prevention (CDS) USA, only in the USA, the cost of falls is expected to reach \$67 billion by the year 2020 \cite{cdc}. Most of the time, elder people live alone in elder care center. They are always vulnerable to fall. Monitoring their activity is very important so that if something bad (such as fall) happens, a staff member or caregiver is informed. Lying on the floor for a long time after the fall may increase the chances of death \cite{RN34}. In recent years, significant work has been done in the field of fall detection. Table 5 gives some applications of fall detection along with the approach, technology, advantages, and disadvantages of different solutions discussed for fall detection. Some of these solutions are based on wearable sensors. Cheng \cite{RN42} proposed a solution for fall detection using tri-axle accelerometer integrated with active RFID tags. The tri-axle accelerometer can provide three directional acceleration values and using this information, the system can classify the human posture into different categories. This study uses a neural network for classification of posture types. A solution for fall detection based on smart-phone was presented in 2017 \cite{RN43}. This technique uses the accelerometer sensor embedded in smart-phone for fall detection. When a fall is detected, a notification, along with the location information, is sent to the pre-defined contacts. Jatesiktat \& Ang   \cite{RN44} proposed a solution based on a wrist-worn device. The wrist-worn device consists of accelerometer and barometer and is worn by the subject all the time. Gia et al. \cite{RN45} presented a solution based on wireless and energy efficient wearable devices. The device has multiple sensors such as accelerometer, gyroscope, magnetometer and temperature sensor. It also has a microcontroller on board, which collects data from these sensors and sends it to a processing unit via a wireless network. Many other solutions have been proposed based on a wearable device for fall recognition \cite{RN46,RN47}.  A major disadvantage in these types of solutions is that carrying a device is not always feasible, especially for elderly people and patients. They may forget about the sensors or may be bothered by wearing a device all the times.

One other approach, which is very popular nowadays, is a device-free approach. Some of the device-free approaches use Wi-Fi for fall detection \cite{RN48,RN49}. Wang et al. \cite{RN50} proposed a solution for fall detection, based on a wireless network. The basic idea of this work is that human activities can affect wireless signals and  CSI  is a good indicator for detecting human activity (fall). A technique proposed by Minvielle et al. \cite{RN51} uses special sensors deployed in the floor for fall detection. Kianoush et al. \cite{RN52} presented a solution based on RF signal using radio devices to detect fall in industrial workplaces.

\begin{normalsize}
\begin{table*}[!t]
\centering
\caption{\label{tab: 4} Approach, technology, advantages, disadvantages and applications of the techniques presented for activities of daily living}
\begin{adjustbox}{width=\textwidth}
\begin{tabular}{|p{2cm}|p{3.2cm}|p{2cm}|p{3.5cm}|p{3.2cm}|}
\hline
\textbf{Approach} & \textbf{Technology}                                    & \textbf{Advantages}        & \textbf{Disadvantages}                                     & \textbf{Applications}                                                \\ \hline
Vision Based      & Camera                                                 & High accuracy              & High cost, complex computation, privacy issue              & \multirow{5}{*}{\begin{minipage}{3.2 cm}Security \& surveillance, \\Smart home, \\Care centers\end{minipage}} \\ \cline{1-4}
Wearable devices  & Accelerometer, temprature sensor, altimeter, gyroscope & Low cost                   & Constraint to carry the device                             &                                                                      \\ \cline{1-4}
Hybrid            & RFID + Wearable device                                 & Low cost                   & Customized harware required, constraint to wear the device &                                                                      \\ \cline{1-4}
Device free       & Motion sensor, proximity sensors, temprature sensor    & Low cost, freedom for user & Environmental interference                                 &                                                                      \\ \cline{1-4}
Object tagged     & Accelerometer, RFID                                    & Low cost                   & Device bound                                               &                                                                      \\ \hline

\end{tabular}
\end{adjustbox}
\end{table*}
\end{normalsize}

Some techniques use passive RFID technology for fall recognition. Wickramasinghe et al. \cite{RN54} proposed the use of passive RFID tags deployed on the floor, for fall recognition. This technique uses tags fitted inside the carpet in a two-dimensional grid and hidden from the users. Unlike their previous work \cite{RN55}, this technique uses binary tag observation information i.e., presence or absence of a tag instead of RSSI which is more vulnerable to environmental noise. Tag observation information is formulated as a binary image i.e., present or absence of a tag when activity happens. This allows the technique to focus on a specific area as a possible location of fall instead of the whole floor because the tags in that area will be marked as unread or absent. When the data from all the tags is received (in form of tag IDs either read or not), it is treated as binary image i.e., some tags will be blocked by the person present while the rest will be read by the reader. The area with the maximum connected region is (where the tags are blocked) selected heuristically as a possible fall region. Only this area is considered for further processing instead of the whole carpet area which significantly reduces the processing cost. Eight features are selected from tag observation information and four different classifiers have been used to classify the activity as fall or not. Different experiments are performed to evaluate the performance of the given technique in a realistic environment. Although the given technique performs better as compared to the previous work of the authors, the proposed system needs to be evaluated for multiple subjects as well as subjects with bags or pets. Also, it is not clear from the paper that how will this technique differentiate between a fall or normal sitting or lying, covering exactly the same number of tags as in a fall. This paper lacks details about the time complexity of the proposed system.

Ruan et al. \cite{RN37} proposed a device-free solution called TagFall, using passive RFID tags which can sense normal activities as well as falls. This technique not only can detect a fall but can also provide information about the direction of fall. TagFall uses the abrupt fluctuation/changes in RSSI caused by falling. For fall detection, TagFall uses Angle Based Outlier Detection method to mine the clustering patterns of RSSI created by normal human activities and detects an anomaly pattern caused by a fall. To detect fall direction, Dynamic Time Warping algorithm is used in which a fixed length data stream is taken and compared with the previously collected profiling data to find the falling direction. After pre-processing, RSSI values are classified into four categories: sitting, lying, standing, and walking. The angle variance of vector pairs formed by the same category is calculated and the upper and lower boundaries of variance are decided. Also, the segmented data streams for falls with different directions, are collected for use in DTW calculations. The basic working principle of this approach is that the angles between different vector pairs from the same activity will differ widely, thus having a high angle variance. Angles between vector pairs from different activity are much smaller. Using this phenomenon, TagFall is able to cluster normal activities and can detect an outlier i.e., fall. Authors have performed extensive experiments to evaluate the performance of TagFall. One of the limitations of this work is that it is designed only for a single resident. This technique is also labor intensive in terms of user profiling and data collection.

\vfill
\subsubsection{Activities of Daily Living}
Recognition of Activities of Daily Living (ADL) is identifying the daily activities in an indoor environment such as a home. These activities include eating, cooking, sleeping, sitting, bathing, dressing, toileting, etc. Recognition of such activities is of great importance for its applications in various areas such as smart homes, surveillance and care centers. A smart home can adapt itself accordingly if it knows the activity of the resident. Recognizing the daily activities of patients or elder people in a caring facility or old homes, can help caregivers to monitor their health and provide the required treatment. Many solutions have been proposed over the past decade to recognize human daily activities. Table 6 presents the applications of daily activity recognition and provide details such as approach, technology, pros and cons of different techniques presented in this section. Some of these techniques use surveillance cameras to capture image or video and then apply computer vision techniques to recognize the activities performed \cite{RN128,RN129}. As mentioned in earlier sections, vision-based techniques have better accuracy but there are many limitations of this approach.

Sensor-based techniques use a different kind of sensors such as accelerometers, motion sensors, pressure sensors, and RFID tags, for recognition of daily activities. Chernbumroong et al. \cite{RN130} proposed a technique based on wrist-worn sensors, for recognition of elder people's activities to support independent living. Three types of sensors are attached to wrist-worn watch of the users which are: accelerometer, temperature sensor, and altimeter. This technique considers both basic ADL (BADL) and instrumental ADL (IADL). BADL includes activities such as grooming, feeding, stairs, dressing, and mobility (walking) while IADL includes activities such as ironing, sweeping, washing dishes and leisure activities (e.g., watching TV). The proposed system achieves an accuracy of more than 90\%. A similar technique was presented by Liu et al. \cite{RN24} for recognition of housekeeping tasks, using accelerometer and gyroscope as wrist-worn sensors.  Wang et al. \cite{RN29} proposed a solution for activity recognition by combining both the RFID system and wearable sensors. They use the RF signals from passive RFID tags connected to the subject's dress. A small reader is also attached to the user's dress, which further extends the coverage area.

\begin{normalsize}
\begin{table*}[!t]
\centering
\caption{\label{tab: 4} Approach, technology, advantages, disadvantages and applications of the techniques presented for ambient assisting living}
\begin{adjustbox}{width=\textwidth}
\begin{tabular}{|p{2cm}|p{3cm}|p{2cm}|p{4cm}|p{3cm}|}
\hline
\textbf{Approach} & \textbf{Technology}                      & \textbf{Advantages} & \textbf{Disadvantages}                        & \textbf{Applications}                                                          \\ \hline
Vision Based      & Camera                                   & High accuracy       & High cost, complex computation, privacy issue & \multirow{3}{*}{\begin{minipage}{3 cm}Elder care center, \\Medication management, \\Exercise management\end{minipage}} \\ \cline{1-4}
Wearable devices  & Inertial sensors, RFID, infrared sesnsor & Low cost            & Constraint to carry the device                &                                                                                \\ \cline{1-4}
Hybrid            & RFID + RF beacons + other sensors        & High accuracy       & High cost, customized hardware required       &                                                                                \\ \hline

\end{tabular}
\end{adjustbox}
\end{table*}
 \end{normalsize}

Some techniques use a hybrid approach by combining both wearable and object-tagged mechanisms. In these techniques, users need to wear a device and objects of daily use are also tagged with different sensors such as accelerometer or RFID. Stikic et al. \cite{RN137} proposed a technique which uses an accelerometer as wrist-worn sensor and objects of daily use are tagged with RFID tags. The authors evaluated their technique in three ways: using data only from the accelerometer, using the data only from RFID tags and using the data from both accelerometer and RFID tags. The results show that the hybrid approach (i.e., combining data from both accelerometer and RFID tags) achieves better results as compared to separate approaches. A similar approach was presented by Hein \& Kirste \cite{RN138} in which the user has to wear a device consisting of an accelerometer, gyroscope, magnetometer, and an RFID antenna. Different objects of daily use are also tagged with RFID tags. The authors evaluated the proposed system in two scenarios: breakfast (preparing and having breakfast, washing the dishes, etc.) and home care with an accuracy of 97\% and 85\%, respectively. Instead of wearing special devices, inertial sensors which are embedded in mobile phones can also be used for recognition of daily life activities \cite{RN132}.

Some techniques use dense sensing and deploy different sensors such as motion sensors, pressure sensor, temperature sensor, and proximity sensor, in the environment \cite{RN133,RN134}. When a user performs any activity in the vicinity of these sensors, relative information can be captured through these sensors which can be used for recognition of activities.

Widely used approach for recognizing ADL is to attach different sensors to objects of daily use and use the interactions of users with these objects to recognize the activity. Different sensors have been used for this purpose but RFID tags and accelerometer are among the most common ones \cite{RN136}. Buettner et al. \cite{RN135} proposed a technique using Wireless Identification and Sensing Platform (WISP) which combines passive RFID tag and accelerometer. Objects of daily use in the kitchen such as cup, bowl, milk-pack, and kettle are tagged with these WISPS and a reader captures the interaction of users with these objects. After collecting the sensor data, HMM is used as an inference engine to infer the activities from the collected data.

\subsubsection{Ambient Assisted Living}
The population is aging around the world because of the low birth rate and increasing life expectancy. According to the Australian Institute of Health and Welfare, 15\% of the Australian population is 65 or over and this number will double by 2056 \cite{cdc}. With the aging population, comes the problem of medical cost and caring of old people. Most of the elder people live alone in their own homes or in elder care facilities. They also need someone to look after them which causes further problems for the workforce. In recent years, considerable research has been done to provide solutions for such problems. Researchers have developed many different technologies to assist humans in their daily lives, under a new paradigm called ambient intelligence. These technologies are called Ambient Assisted Living (AAL) tools and are helping people with issues such as remote monitoring, medication management, medication reminder, exercise management, and independent living. Over the last decade, many solutions have been proposed under the umbrella of AAL to support independent living of the elder people \cite{RN139,RN140}. Table 7 gives some details (approach, technology, advantages, disadvantages) about different solutions discussed in this section along with some applications of ambient assisted living. Some of these solutions are vision-based and use surveillance camera to capture the information about the activities of the residents \cite{RN141}. As discussed before, vision-based systems have many issues.

A number of other solutions have been proposed using different types of sensors. These sensors are used in two ways: as wearable and attached to objects of daily use. Zhu \& Sheng \cite{RN142} proposed a multi-sensor technique for recognition of daily activities in robot-assisted living. Two inertial sensors are attached to the body of the user, one on the waist and other on the foot. Sensors are connected to a PDA which transfers the sensor data (angular velocity and the acceleration) to a desktop computer through Wi-Fi for processing. A set of neural networks classify the data into three categories: transitional, stationary, and cyclic. The output from the neural networks is fed into a fusion module which further categorizes them as zero displacement activity, transitional activity, and strong displacement activity. Zero displacement activities are further classified into sitting or standing while transitional activities are classified into standing-to-sitting or setting-to-standing by using a heuristic discrimination module. Strong displacement activities are further classified by applying the HMM algorithm. This approach is tested in a real-life scenario and achieved around 98\% accuracy.

\begin{normalsize}
\begin{table*}[!t]
\centering
\caption{Comparison of different approaches for recognizing action based activities.\newline Symbols used: D= device-free, W= wearable, H= hybrid, Y= yes, N= no, -- = Not Available}
\label{my-label}
\begin{adjustbox}{width=\textwidth}
\begin{tabular}{|p{2.5cm} |p{0.7cm}|p{0.7cm} |p{3.5cm}|p{3cm}|p{3cm}|p{0.7cm}|}  

\hline

\multicolumn{7}{|l|}{M1 = Approach, M2 = Technology, M3 = Information Type, M4 = ML Algorithm, M5 = Supervised/Unsupervised}                                                                             \\ \hline
\textbf{Category}                      & \textbf{Paper} & \textbf{M1} & \textbf{M2}                                   & \textbf{M3}                & \textbf{M4}                                           & \textbf{M5} \\ \hline
\multirow{4}{*}{\begin{minipage}{2.5 cm}Gesture Recognition\end{minipage}}   & \cite{RN74}             & D           & RFID                                          & Tag ID                     & Fisher's Linear Discriminant                          & Y           \\ \cline{2-7} 
                                       & \cite{RN75}             & D           & RFID                                          & RSSI                       & K-Means Clustering                                    & N           \\ \cline{2-7} 
                                       & \cite{RN76}            & D           & RFID                                          & RSSI, Phase, Tag ID        & -                                                     & N           \\ \cline{2-7} 
                                       & \cite{RN77}              & D           & RFID                                          & Phase Values               & N-DTW, Weighted Matching Algorithm                    & Y           \\ \hline
\multirow{5}{*}{Fall Detection}        & \cite{RN73}              & D           & RFID                                          & RSSI                       & KNN                                                   & Y           \\ \cline{2-7} 
                                       & \cite{RN50}            & D           & Wi-Fi                                         & CSI                        & Random Forst                                          & Y           \\ \cline{2-7} 
                                       & \cite{RN51}             & D           & Piezoelectric Polymer Sesnor                  & Electric Signal            & Random Forest                                         & Y           \\ \cline{2-7} 
                                       & \cite{RN52}             & D           & Radio Frequency                               & RSSI                       & HMM                                                   & Y           \\ \cline{2-7} 
                                       & \cite{RN54}             & D           & RFID                                          & Tag IDs                    & NSVM                                                  & Y           \\ \hline
\multirow{5}{*}{\begin{minipage}{2.5 cm}Posture Recognition\end{minipage}}   & \cite{RN13}             & D           & RFID                                          & RSSI                       & DPGMM based HMM                                       & Y           \\ \cline{2-7} 
                                       & \cite{RN19}             & D           & RFID                                          & RSSI                       & SVM                                                   & Y           \\ \cline{2-7} 
                                       & \cite{RN117}              & D           & RFID                                          & Phase, RSSI                & -                                                     & Y           \\ \cline{2-7} 
                                       & \cite{RN87}              & D           & Radar                                         & RF Samples                 & SVM,NB, KNN, RF, Logistic Regression                  & Y           \\ \cline{2-7} 
                                       & \cite{RN22}            & W           & CRFID                                         & RSSI                       & NB, CRF, RF, LSVM, NSVM                               & Y           \\ \hline
\multirow{4}{*}{\begin{minipage}{2.5 cm}Behaviour Recognition\end{minipage}} &  \cite{RN93}             & D           & Kinect Sensor                                 & Silhouettes                & SVM, K-NN, LDC                                        & Y           \\ \cline{2-7} 
                                       & \cite{RN94}             & D           & Wi-Fi                                         & CSI                        & D-Tree, Simple Logistic Regre                         & Y           \\ \cline{2-7} 
                                       & \cite{RN96}            & D           & RFID                                          & Phase, Doppler's Effect    & Iterative Clustering Algorithm With Cosine Similarity & N           \\ \cline{2-7} 
                                       & \cite{RN97}             & D           & RFID                                          & Phase Values               & -                                                     & N           \\ \hline
\multirow{5}{*}{ADL}                   & \cite{RN137}          & W           & Hybrid(RFID+Accelerometer)                    & Accelerometer data, Tag ID & NB, HMM, Joint Boosting                               & Y           \\ \cline{2-7} 
                                       & \cite{RN135}             & D           & Hybrid(RFID+Accelerometer)                    & Accelerometer data, Tag ID & HMM                                                   & Y           \\ \cline{2-7} 
                                       & \cite{RN138}           & W           & Hybrid(RFID+Inertial Sensor)                  & Data from IMU, Tag ID      & HMM, Weka C4.5                                        & Y           \\ \cline{2-7} 
                                       & \cite{RN130}           & W           & Hybrid(Accelerometer,Altimeter, Temp. Sensor) & Data from All Sensors      & SVM                                                   & Y           \\ \cline{2-7} 
                                       & \cite{RN24}              & W           & Accelerometer, Gyroscope                      & Sensor Readings            & NB, D-Tree, KNN, SVM                                  & Y           \\ \hline
\multirow{5}{*}{AAL}                   & \cite{RN143}            & D           & Infrared Sensors                              & Sensor Readings            & Polya's urn                                           & Y           \\ \cline{2-7} 
                                       & \cite{RN142}           & W           & Inertial Sesnors                              & Sensor Readings            & Neural Network, HMM                                   & Y           \\ \cline{2-7} 
                                       & \cite{RN144}            & D           & RFID                                          & RSSI, Phase, Tag ID        & Information Gain Algorithm                            & Y           \\ \cline{2-7} 
                                       & \cite{RN145}            & W           & RFID, Pressure Sensor                         & RSSI, Sensor Reading       & -                                                     & -           \\ \cline{2-7} 
                                       & \cite{adame2018cuidats}            & H           & RFID, WSN                                     & Data from All Sensors      & -                                                     & -           \\ \hline

\end{tabular}
\end{adjustbox}
\end{table*}
\end{normalsize}

\begin{normalsize}
\begin{table*}[!t]
\centering
\caption{Comparison of different approaches for recognizing action based activities.\newline Symbols used: D= device-free, W= wearable, H= hybrid, Y= yes, N= no, -- = Not Available }
\label{my-label}
\begin{adjustbox}{width=\textwidth}
\begin{tabular}{|p{2.5 cm} |p{1cm}|p{5cm} |p{1cm}|p{1cm}|p{1cm}|p{1cm}|}  

\hline
\multicolumn{7}{|l|}{M6 = Application, M7 = Cost, M8 = Accuracy, M9 = Latency, M10 = Real-Time}                  \\ \hline
\textbf{Category}                      & \textbf{Paper} & \textbf{M6}                                & \textbf{M7} & \textbf{M8} & \textbf{M9} & \textbf{M10} \\ \hline
\multirow{4}{*}{\begin{minipage}{2.5 cm}Gesture Recognition\end{minipage}}   & \cite{RN74}             & 12 Gestures                                & Low         & High        & -           & N            \\ \cline{2-7} 
                                       & \cite{RN75}            & 2 Gestures                                 & Low         & High        & 2.95 s      & Y            \\ \cline{2-7} 
                                       & \cite{RN76}             & 7 Gestures                                 & Low         & High        & 0.1 s       & Y            \\ \cline{2-7} 
                                       & \cite{RN77}             & 6 Gestures                                 & Low         & High        & -           & -            \\ \hline
\multirow{5}{*}{Fall Detection}        & \cite{RN73}             & Postures, Fall Detection Direction of Fall & Low         & High        & -           & -            \\ \cline{2-7} 
                                       & \cite{RN50}             & Postures, Fall Detection                   & Medium      & Medium      & -           & -            \\ \cline{2-7} 
                                       & \cite{RN51}             & Fall Detection                             & Low         & High        & -           & -            \\ \cline{2-7} 
                                       & \cite{RN52}             & Localization and Fall Detection            & Medium      & High        & -           & Y            \\ \cline{2-7} 
                                       & \cite{RN54}               & Fall Detection                             & Low         & High        & 1.5 s       & Y            \\ \hline
\multirow{5}{*}{\begin{minipage}{2.5 cm}Posture Recognition\end{minipage}}   & \cite{RN13}             & Postures, Posture Transition                & Low         & High        & 3.5 s       & Y            \\ \cline{2-7} 
                                       & \cite{RN19}             & Postures, Actions                          & Low         & High        & 4.5 s       & Y            \\ \cline{2-7} 
                                       & \cite{RN117}             & Postures, Gestures                         & Low         & Medium      & -           & -            \\ \cline{2-7} 
                                       & \cite{RN87}              & Office Desk \& Checkout Counter Activities & Medium      & High        & -           & N            \\ \cline{2-7} 
                                       & \cite{RN22}             & Bed-exit, Chair-exit, walking              & Medium      & High        & -           & Y            \\ \hline
\multirow{4}{*}{\begin{minipage}{2.5 cm}Behaviour Recognition\end{minipage}} & \cite{RN93}             & 6 Actions/Activities                       & High        & Medium      & -           & Y            \\ \cline{2-7} 
                                       & \cite{RN94}             & 3 Coarse-grained Activities                & Medium      & High        & -           & -            \\ \cline{2-7} 
                                       & \cite{RN96}            & 3 Types of Behaviour                       & Low         & High        & -           & N            \\ \cline{2-7} 
                                       & \cite{RN97}            & 3 Types of Behaviour                       & Low         & High        & -           & N            \\ \hline
\multirow{5}{*}{ADL}                   & \cite{RN137}            & 10 Housekeeping Activities                 & Medium      & Medium      & -           & -            \\ \cline{2-7} 
                                       & \cite{RN135}             & 14 Daily Life Activities                   & Medium      & High        & -           & -            \\ \cline{2-7} 
                                       & \cite{RN138}            & 19 Daily Life Activities                   & Medium      & Medium      & -           & -            \\ \cline{2-7} 
                                       & \cite{RN130}            & 11 Daily Life Activities                   & Medium      & High        & -           & -            \\ \cline{2-7} 
                                       & \cite{RN24}              & 12 Daily Life Activities                   & Medium      & High        & -           & N            \\ \hline
\multirow{5}{*}{AAL}                   & \cite{RN143}            & Monitoring Daily Activities                & Low         & Low         & -           & -            \\ \cline{2-7} 
                                       & \cite{RN142}            & 5 Gestures, 4 Postures                     & Low         & High        & -           & -            \\ \cline{2-7} 
                                       & \cite{RN144}            & Interaction with Objects                   & Low         & Medium      & -           & Y            \\ \cline{2-7} 
                                       & \cite{RN145}            & Movement Tracking                          & Low         & Medium      & -           & -            \\ \cline{2-7} 
                                       & \cite{adame2018cuidats}             & Tracking, Localization, Status Monitoring  & High        & Low         & -           & Y            \\ \hline

\end{tabular}
\end{adjustbox}
\end{table*}
\end{normalsize}

Raad et al. \cite{RN145} presented an RFID-based system for monitoring the activities of Alzheimer's patient at home. The basic idea of this work is to track the movement of a user from one room to another and to report any abnormal situation (e.g., staying in the washroom for a longer time). The user has to wear a passive RFID tag around the ankle because the ankle is the relatively stable position in the body. To enhance the system efficiency, two pressure mate sensors are deployed on either side of the door to detect whether a user is coming inside or going outside. When a movement is detected by the sensors, the system triggers the reader to energize the tags and collects the data for detecting the location of the user. The proposed system is tested in a lab environment and achieved 88\% accuracy. An issue with this solution is that wearing a device all the time is not a good choice, especially when it comes to elder people.

Many other solutions have been proposed using dense sensing approach. Franco et al. \cite{RN143} proposed a technique for telemonitoring of elder people using dense sensing. The main objective of this study is to detect the nycthemeral shift in the daily routines of the older people which can help in early detection of dementia-related diseases. Infrared sensors are deployed in different locations of a flat (test facility) to capture the information about the daily activities of the resident. A total of eight months of data is recorded. A random process technique called Polya's urns is used to analyze the recorded data. Prada et al. \cite{RN144} presented a method called Weighted Information Gain (wIG) to detect user-object interaction for assisting independent living. They use RFID technology i.e., RFID tags are attached to different objects (e.g., books). For proper training of the proposed system, light dependent resistors are used to represent the presence or absence of an object (e.g., book in a shelf). The wIG classifier uses the Information Gain algorithm to classify the RFID events as static or interacted. Experimental evaluation shows that the proposed system achieves better accuracy as compared to other similar approaches.

Although RFID technology provides a better solution for autonomous identification and tracking of objects, the range of RFID is an issue. The detection range of RFID is a few meters. To tackle this issue, many researchers have tried to merge RFID technology with other technologies such as Wireless Sensor Network (WSN). One such system called CUIDATS is proposed by Adame et al.\cite{adame2018cuidats}, for monitoring in a smart healthcare facility. The proposed system can track patients and assets and can also monitor vital signs (e.g., temperature, fall alert, pulse). CUIDATS is a hybrid solution in which RF beacons and RF readers are deployed in the environment while patients need to wear a wristband which consists of different sensors and RF transmitter. Assets are tagged with passive and active RFID tags. RFID reader and RF beacon are integrated into a single compact device which can collect the data and transfer it to the WSN. CUIDATS uses weighted centroid algorithm for tracking patients with wristband and accelerometer readings are used for fall detection. The proposed system is evaluated in a real-life environment (a hospital) and different tests are performed with reasonable accuracy.  
\vfill

\subsubsection*{\textbf{Summary}}
A summary of the work presented for action based activities is given in Table 8 and 9. As seen from the tables, most of the solutions use device-free approaches. Some of the solutions also use wearable approach. Substantial works have been done in the areas of gesture recognition, posture recognition, using RFID technology. Many of the solutions especially in the areas of AAL and ADL, are sensor-based and are using sensors like accelerometer, proximity sensor, and other sensors. The accelerometer is the most common sensor used in the field of human activity recognition. 

Because of its device-free nature and easy deployment, RF technology is finding its usage in many fields. A significant amount of work using RF technology can be found in HAR. Many solutions have been presented using RF technology, especially Wi-Fi. Wi-Fi is used as a device-free approach for solutions in behavior recognition, fall detection, and people counting. Due to the fact that Wi-Fi is present almost everywhere nowadays, researchers are using this technology for many applications in various fields and it is providing good results. 

One approach which is becoming very common nowadays is the fusion approach. Instead of using a single technology like accelerometer or RFID, researchers are now using the hybrid approach by combining multiple technologies. One example of such an approach is combining wearable accelerometer sensor with RFID tags attached to objects of daily use. The hybrid approach has the advantages of both approaches.

Machine learning plays an important role in activity recognition. Information can be collected through various approaches and technologies but after that, it is the job of machine learning algorithm to infer/recognize the activity. Some of the most common machine learning algorithms which are used in human activity recognition are; SVM, KNN, Random Forest, Naive Bayes, and HMM. Feature selection is also an important part before applying a machine learning algorithm. A good set of features can give better results.

\vfill

\subsection{Motion Based Activities}
\label{sec:motion-based}
 These types of activities are related to the motion of human being. Activities are not only those which are related to performing any specific action but presence or absence, motion detection, etc., in an area under observation can also be an activity. Recognizing motion based activities is very useful, especially in surveillance and security. In this section, we provide an overview of motion based activities.

\begin{figure}[hbt!]
	\centering
	\includegraphics[width= \linewidth]{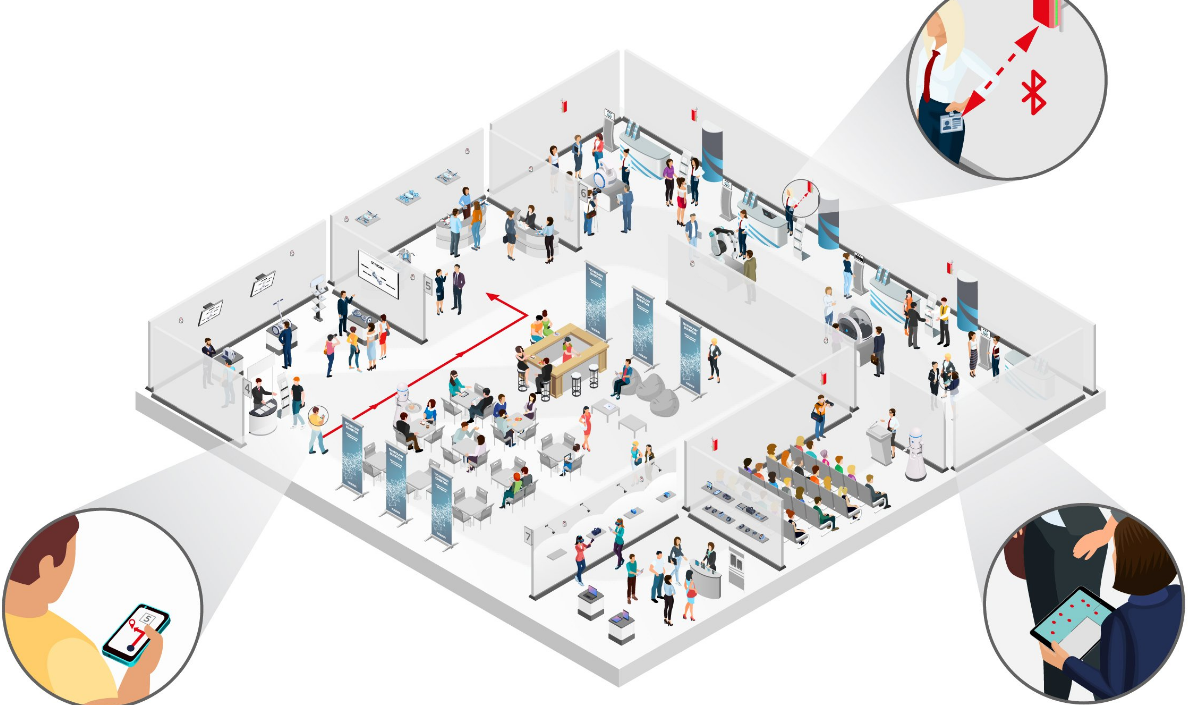}
	\caption{Examples of motion based activities (e.g., path tracking, asset tracking, motion detection in office). \protect\footnotemark}
	\label{fig1}
\end{figure}

\footnotetext{Source: https://www.infsoft.com/examples-of-use/articleid/234/indoor-navigation-and-staff-tracking-at-a-trade-fair}

\subsubsection{Tracking}
Tracking is one of the important sub-areas in human activity recognition. In an outdoor environment, tracking can be easily done using Global Positioning System (GPS) but GPS is not applicable in indoor environments. Tracking has many uses in various applications such as augmented reality, room occupancy detection, and indoor navigation. Due to its increasing importance, significant work has been done in this area. Table 10 gives some applications and details such as approach, technology, pros and cons for the techniques presented in this section.  Existing research work in this area can be divided into device-bound \cite{RN98} and device-free approaches. One of the limitations of the device-bound approaches is that the subject is required to carry a device or tag. But carrying a device or tag is not possible in all the cases, for example, in cases of animal tracking and unknown subjects. In device-free approaches, users are free from carrying any devices. One such example is the use of Wi-Fi signals for tracking the motion of humans \cite{RN100,RN101}.

One approach for device-free tracking is to use passive RFID technology \cite{RN7,RN11,RN5,RN103,RN104,RN99,RN106}. But the main challenge in this approach is the interference from the environment, which affects the accuracy of the solution \cite{RN99}. 

\begin{normalsize}
\begin{table*}[hbt!]
\centering
\caption{\label{tab: 4} Approach, technology, advantages, disadvantages and applications of the techniques presented for tracking}
\begin{adjustbox}{width=\textwidth}
\begin{tabular}{|p{1.7 cm}|p{3 cm}|p{3cm}|p{2.5 cm}|p{3.5cm}|}
\hline
\textbf{Approach}            & \textbf{Technology}      & \textbf{Advantages}                      & \textbf{Disadvantages}        & \textbf{Applications}                                                          \\ \hline
Wearable device              & Accelerometer, gyroscope & Low cost                                 & Constraint to wear the device & \multirow{4}{*}{\begin{minipage}{3.5 cm}Supply chain management, \\indoor navigation, \\augmented reality\end{minipage}} \\ \cline{1-4}
\multirow{3}{*}{Device free} & Wi-Fi                    & Low cost, COTS available                 & Environmental interference    &                                                                                \\ \cline{2-4}
                             & RFID (passive)           & Low cost, COTS available, passive        & Environmental interference    &                                                                                \\ \cline{2-4}
                             & RFID(passive + active)   & Low cost, COTS available, high accuracy, & Active tags need battery      &                                                                                \\ \hline

\end{tabular}
\end{adjustbox}
\end{table*}
\end{normalsize} 

TASA \cite{RN7} proposed a device-free RFID-based approach for location sensing and frequent route detection. This approach uses the RSSI values of the tags, arranged in arrays and deployed in the locality where the object will move, to find the frequent trajectories. To improve accuracy, active tags are used as referenced tags with known locations. This technique model the whole tag array in a coordinate system in which each tag represents a specific coordinate value with respect to a reference tag. The entire process is divided into two phases: location sensing and frequent route detection. In the first phase, RSSI values of only the affected tags (whose value is greater than a threshold $\lambda$) are taken into account and are stored in a database after sorting chronologically.  This approach significantly reduces the memory requirement. Also, multiple readers are used and their readings are sorted in chronological order. Outliers are removed using the idea that only those tags are considered as affected if their neighbor tags are also affected otherwise these will be outliers. For locating objects along the trajectory, TASA makes use of the active tags deployed in critical positions along with passive tags. It is because active tags are more responsive to changes as compared to passive tags. At the end of the first phase, raw RSSI data has been converted to a set of different routes in chronological order. Phase two is activity sensing in which frequent routes are detected by using a two-step approach; frequent route set discovery with minimum support and online detection of frequent routes. TASA uses modified versions of Apriori \cite{RN9} and FPGrowth \cite{RN10} algorithm called as iApriori and iFPGrowth for detecting frequent trajectories. TASA is also capable of tracking multiple objects simultaneously with the help of active reference tags. Experimental evaluations show that TASA can perform better in different scenarios and even when the tags and readers are hidden from the subjects. TASA can accurately track single or multiple objects. However, there are some limitations of this approach. The accuracy is low when used to track more than two objects simultaneously. Also, active tags require battery maintenance which is not feasible in some situations. Placement of active tags is another issue. This approach is using parameters ($\lambda$ and minimum support for iApriori) and tuning of parameters is always a challenging task. As shown by results, the performance is sensitive to parameter tuning and tags placement.

\begin{normalsize}
\begin{table*}[hbt!]
\centering
\caption{\label{tab: 4} Approach, technology, advantages, disadvantages and applications of the techniques presented for motion detection}
\begin{adjustbox}{width=\textwidth}
\begin{tabular}{|p{2cm}|p{1.5cm}|p{3cm}|p{3.5cm}|p{3cm}|}
\hline

\textbf{Approach}                & \textbf{Technology} & \textbf{Advantages}               & \textbf{Disadvantages}                        & \textbf{Applications}                                   \\ \hline
Vision based                     & Camera              & High accuracy                     & High cost, complex computation, privacy issue & \multirow{3}{*}{\begin{minipage}{3 cm}Security \& surveillance, \\smart homes\end{minipage}} \\ \cline{1-4}
\multirow{2}{*}{Radio frequency} & Wi-Fi               & Low cost, COTS available          & Environmental interference                    &                                                         \\ \cline{2-4}
                                 & RFID                & Low cost, COTS available, passive & Environmental interference                    &                                                         \\ \hline

\end{tabular}
\end{adjustbox}
\end{table*}
\end{normalsize}

Liu et al., \cite{RN11} presented a device-free, RFID-based, approach for mining frequent trajectories. Active RFID tags along with readers are deployed. When an object moves around in this area, the tags along the path of the movement will be disturbed and the RSSI information from these tags is used for detecting a trajectory. Before the data collections, two base values are calculated for the tags in the absence of any objects. These values are neutral values of a tag which is the expected signal strength and sensitivity of the tag. When data (signal strength indicator) is collected from the tags, neutral value and RF values are used to find the interfered tags. If a tag is interfered, the signal strength value is replaced by one and if the tag is not interfered, it is replaced by zero. In this way, the data is transformed into a binary form. After the pre-processing stage, the next phase is to mine frequent trajectories. This task is done in two steps: training and monitoring. In the training phase, data is collected from tags for a certain period of time and this data is used to find frequent trajectories which are modeled as normal activities. In the monitoring phase, activity is detected and is compared with the frequent trajectories. The activity is considered normal or suspicious based on the comparison results. This work focuses on training phase to mine frequent trajectories because in the author's opinion, trajectory matching is the same problem as approximate sequence matching and there are many solutions for this problem. Some of the issues related with RFID tags are: not every tag along the trajectory may be detected, detected tags may not be able to accurately reflect the activity's order, and it is also possible that multiple tags are inferred by the activity in a given period. To tackle these issues, this work identifies the border between interfered and non-interfered tags for an activity. After the borders are detected, the possible positions of the objects are identified using the spatial map of the tags which is known in advance. Instead of the exact location, the ranges are located where the object possibly exists. One issue is that there may be more than one objects present in that area e.g., one object may be hidden behind the other. To handle this issue, two approaches are adopted. One option is to deploy multiple readers because the hidden object may be detected by at least one of the readers. The second option is to do fault-tolerant mining e.g., the hidden object may show up in next time periods. Mining algorithm consists of two parts. In the first part, frequent positions of the object are identified. In the second part, frequent trajectory segments are calculated. Starting with the short segment, frequent segments are extended using a depth-first search. The authors have performed different experiments to evaluate the performance of this technique in various scenarios. Although this technique performs well in many scenarios, there are some issues. Parameter tuning is required which is not an easy task. Active tags are used which require maintenance for battery replacement. This work does not provide any details about the tag or reader placement.

TagTrack \cite{RN5} is another device-free technique which uses passive RFID tags for tracking. The basic idea of this work is that RSSI shows different patterns when a person is present or absent in a given RSS field. When a person moves through different regions in a given RSS field, the RSSI pattern changes accordingly. This work focuses on two main problems: localization of a stationary object and tracking of a moving object. Localization is considered as a classification problem and different classification techniques are applied to locate a stationary object. TagTrack proposes two techniques: GMM-based HMM and kNN-based HMM, to track the moving object by learning the underlying patterns in different locations. Experimental results show that the given system can perform better for localization but the performance for tracking a moving object is poor.

Tadar is a system proposed by Yang et al. \cite{RN103} which can track moving objects, even beyond the wall. Passive RFID tags are attached to outer walls with a reader fixed in line of sight. Basic idea behind Tadar is that tags receive the signal from the reader via two paths; directly from the reader and reader's signal reflected by another object and then received by the tag. Tadar exploits this reflected signal for tracking the object. HMM is used for object tracking.  The proposed system has some problems such as direction dependency and vulnerability to reflective (metallic) objects. The detection range of the system is low (around 4 meters) for a concrete wall but most of the buildings use concrete walls. Also, the given system can track only one moving object and is not applicable for multiple objects.

Han et al. \cite{RN104} presented a device-free RFID-based technique called Twins, for tracking and motion detection. This technique uses a phenomenon called the critical state, which is caused by the interference of different passive tags. The working principle of Twin is based on the critical state caused by the coupling effect among passive RFID tags. The idea is when two passive tags (e.g., A and B) are placed together at a certain distance, with the same antenna, one of the tags (B) become unreadable. It's because of the shadow effect from the other tag (A) and because of this effect, tag B's antenna will receive a very weak signal from the reader and therefore, will not respond to the reader. But if an object (human) pass close to the twins, some of the RF waves get reflected or refracted. Because of this disturbance, tag B receives enough energy to break the critical stage and becomes readable again. In this way, a moving object can be detected.

\subsubsection{Motion Detection}
Motion detection is a process to detect the presence of any moving entity in an area of interest. Motion detection is of great importance due to its application in various areas such as surveillance and security, smart homes, and health monitoring. A smart home is smart because it can adjust its environment according to the user's activity. The first and most basic thing for that is to know about the presence or absence of a resident. In security and surveillance, intrusion detection is very important and one of the basic tasks, which is basically detecting the presence or motion of outsiders. Motion detection also plays a key role in the field of health and remote monitoring of patients especially elder people. Different approaches have been used to provide solutions for motion detection. Details such as technology, approach, advantages, and disadvantages for solutions discussed in this section are given in Table 11 along with the applications of motion detection. Some of these solutions are vision-based and use surveillance cameras \cite{RN107}. Some techniques use wearable sensors attached to the subject, for motion and presence detection. But this approach requires a sensor or device to be worn by a subject which is not possible in some cases, for example, unknown intruders or animals. 

\begin{normalsize}
\begin{table*}[hbt!]
\centering
\caption{\label{tab: 4} Approach, technology, advantages, disadvantages and applications of the techniques presented for people counting}
\begin{adjustbox}{width=\textwidth}
\begin{tabular}{|p{2cm}|p{2cm}|p{2.9cm}|p{3.3cm}|p{2.8cm}|}
\hline

\textbf{Approach}            & \textbf{Technology}    & \textbf{Advantages}               & \textbf{Disadvantages}                              & \textbf{Applications}                                                                   \\ \hline
Vision based                 & Camera                 & High accuracy                     & High cost, complex computation, privacy issue       & \multirow{6}{*}{\begin{minipage}{3 cm}Crowd management, \\shopping malls, \\public gatherings, \\Smart environments\end{minipage}} \\ \cline{1-4}
Depth sensor                 & Kinect, infrared laser & High accuracy                     & High cost, privacy issue                            &                                                                                         \\ \cline{1-4}
\multirow{2}{*}{Gate based}  & Turnstile gate         & High accuracy                     & Require the user to pass through a specific area (gate) &                                                                                         \\ \cline{2-4}
                             & Laser beam             & High accuracy                     & Require user to pass through a specific area (gate) &                                                                                         \\ \cline{1-4}
\multirow{2}{*}{Device free} & Wi-Fi                  & Low cost, COTS available          & Cannot work if number of people increase            &                                                                                         \\ \cline{2-4}
                             & RFID                   & Low cost, COTS available, passive & Cannot work if number of people increase            &                                                                                         \\ \hline

\end{tabular}
\end{adjustbox}
\end{table*}
\end{normalsize}

One class of solutions adopt a device-free approach for motion detection\cite{RN108,RN109}. With the widespread deployment of Wi-Fi network, many solutions have been proposed using Wi-Fi signals for motion detection. These solutions exploit the changes induced by human motion in RF signal for detecting motion or presence of the human subject. Some examples of such solutions are FIMD \cite{RN110}, RoMD \cite{RN111}, and MoSense \cite{RN112}.

Zhao et al. \cite{RN113} proposed a technique called EMoD which is not only capable of locating a moving object but can also provide information about the direction of movement. EMoD is a device-free technique and uses passive RFID tags. Passive RFID tags are deployed in pairs (twins) at different points in an area under observation. The working principle of EMoD is the same as in Twins \cite{RN104} i.e., critical state of the tags.

A device-free technique based on passive RFID technology called RF-HMS was presented by Wang et al.\cite{RN115}. Like Tadar \cite{RN103}, this technique uses RFID tags for seeing through the wall. By deploying passive tags on the outer side of the wall, RF-HMS can detect the presence of a stationary human, moving human, and also the direction of movement. RF-HMS does not require prior learning of the empty room's environment. RF-HMS characterizes each tag's multi-path propagation by channel transfer function using phase and RSSI measurements. It eliminates the noise and reflection from static entities such as furniture and walls, by dividing channel transfer function, learned beforehand for each tag, irrespective of the presence or absence of a human in the room. Passive RFID tags are grouped in the form of an array to improve the sensing performance. Reflections from walls, indoor furniture and various parts of the human body are captured by the tag arrays and are combined by RF-HMS into a reinforced result. Then phase shifts can be extracted to detect the presence or absence of a human in the room. This solution can provide information about only two directions i.e., towards the tag or away from the tag and cannot detect motion in other directions such as left or right. Also, this technique requires calibration of threshold values which is always a challenging task. The proposed solution needs to be evaluated for concrete walls as most of the buildings have concrete walls.

\subsubsection{People Counting}
People counting means counting or estimating the number of people in a specific area, which can be a closed environment or an open area \cite{RN87}. People counting is of great importance in various people-centric IoT applications like smart homes, elder care centers, and traffic management. This process has many applications both in normal and critical situations \cite{RN87}. Some examples of critical situations are crowd control in huge festivals, public gathering, religious festivals, music concerts, sports stadium, etc. It is important to know the number of people attending a specific event so that the required arrangements can be made. An example of non-critical situations in which people counting has many applications is, counting the number of people visiting a specific facility (e.g., museum, retail store, train station, shopping mall, restaurant, art galleries or a library). Through the people counting system, the number of customers waiting in a queue or waiting room can be estimated. People waiting to be served is common in many facilities such as airport, restaurants, theme parks, and hospitals. It is very important for the managers to know the number of people waiting or visiting their facility and can help them provide better service to the customers, thus, enhancing their business. Through the people counting system, the trend of visiting people can be found. For example, a mall may have more visitors on Thursday as compared to other days, a library may have more visitors on weekdays as compared to weekends. Finding these trends can help business owners to plan accordingly and provide better arrangements to the customers. People counting can be of two types: crowd counting in an area and counting the number of people going in or out of a specific closed environment. Different solutions have been proposed to solve the problem of people counting. Table 12 gives some applications of people counting along with some details for different techniques presented for people counting. These can be categorized as image-based and non-image-based \cite{RN125}. Image-based techniques use cameras to capture an image or video of the area under surveillance and then analyze the image or video to find the number of people presents \cite{RN121,RN122}. Some techniques use depth cameras and infrared lasers instead of traditional cameras \cite{RN123,RN124}.

Non-image-based techniques use binary sensors, mechanical barriers and wireless signals for people counting. Mechanical barrier based techniques use a turnstile gate, which allows only one person at a time, to pass through the gate. This allows counting the number of people passing the gate. Binary sensor-based solution use break-beam sensors such as infrared or laser beam, deployed on a one-way gate \cite{RN127}. When a person passes by the gate, it causes the beam to break allowing to count the number of people passing by the gate. A major problem with this type of solutions is that they require the subject to pass through a specific area (gate) which is not feasible in many situations such as crowd present in an exhibition.

Some solutions use wireless signals (such as Wi-Fi and RF), which is a more economical and practical approach. These solutions do not affect the privacy of people and can use existing infrastructure such as commodity Wi-Fi. Received signal strength of the wireless signal is an indicator of the signal when it propagates through a region. RSS is sensitive to the number of people present in a specific environment and can provide information for finding the number of people. Cheng \& Chang \cite{RN125} proposed a device-free technique for counting the number of people in an indoor environment using Wi-Fi channel state information. They use a transmitter (Wi-Fi access point) and a receiver for collecting the RSS values of the signals. Deep Neural Network model is used in this technique which is trained offline on CSI data for the different number of people present in the room and then tested online.  The proposed system is robust to location variability of the people inside a facility.

With the recent popularity of RFID technology and a decrease in the cost of RFID tags, this technology has found its place in various fields. Solutions to different problems are using RFID technology. One such example is using passive RFID technology for counting the number of people. R\# \cite{RN126} is a device-free technique, which uses passive RFID tags to estimate the number of people present in a facility. The basic idea of this work is that variance in the RSS values of backscattered RF signal change according to the number of people present in the environment. Passive RFID tags are deployed in the area under consideration and RSS is captured by a reader when the different number of people are present in the region. The proposed solution provides an easy and cost-effective technique for counting the number of people and achieves good results as shown by the experiments. A limitation of this work is that it cannot count more than ten people. Also, the performance is poor when people are walking at relatively high speed. This work has ignored the issues like time complexity, the effect of the surrounding (e.g., metallic objects), the effect of multi-height participants, etc.

\begin{normalsize}
\begin{table*}[!t]
\centering
\caption{Comparison of different approaches for recognizing motion based activities.\newline Symbols used: D= device-free, W= wearable, H= hybrid, Y= yes, N= no, -- = Not Available}
\label{my-label}
\begin{adjustbox}{width= \linewidth}
\normalsize
\begin{tabular}{|p{3cm} |p{1.5cm}|p{1cm} |p{2.5cm}|p{4cm}|p{4cm}|p{0.5cm}|}  

\hline
\multicolumn{7}{|l|}{M1= Approach ,M2= Technology, M3= Information Type, M4= ML Algorithm , M5= Supervised/Unsupervised} \\ \hline
\textbf{Category}                       & \textbf{Paper}      & \textbf{M1}      & \textbf{M2}          & \textbf{M3}                  & \textbf{M4}                       & \textbf{M5}      \\ \hline
\multirow{5}{*}{Tracking}               & \cite{RN7}                  & D                & RFID                 & RSSI                         & Apriori, FPGrowth                 & -                \\ \cline{2-7} 
                                        & \cite{RN11}                  & D                & RFID                 & RSSI                         & -                                 & Y                \\ \cline{2-7} 
                                        & \cite{RN5}                  & D                & RFID                 & RSSI                         & GMM based HMM, kNN based HMM      & Y                \\ \cline{2-7} 
                                        & \cite{RN103}                  & D                & RFID                 & RSSI, Phase                  & HMM                               & Y                \\ \cline{2-7} 
                                        & \cite{RN104}                  & D                & RFID                 & RSSI                         & KNN                               & Y                \\ \hline
\multirow{5}{*}{Motion Detection}       & \cite{RN111}                  & D                & RF                   & CSI                          & -                                 & -                \\ \cline{2-7} 
                                        & \cite{RN113}                  & D                & RFID                 & Critical Power of Tags       & -                                 & Y                \\ \cline{2-7} 
                                        & \cite{RN109}                  & D                & Sensor Fusion        & Raw Signal from Sensors      & Mean-shift Clustering             & Y                \\ \cline{2-7} 
                                        & \cite{RN112}                  & D                & RF                   & CSI                          & -                                 & Y                \\ \cline{2-7} 
                                        & \cite{RN115}                  & D                & RFID                 & RSSI, Phase                  & -                                 & NA               \\ \hline
\multirow{4}{*}{People Counting}        & \cite{RN126}                  & D                & RFID                 & RSSI                         & NB                                & Y                \\ \cline{2-7} 
                                        & \cite{RN123}                  & D                & Kinect Sensor        & Depth Image                  & -                                 & -                \\ \cline{2-7} 
                                        & \cite{RN125}                  & D                & Wi-Fi                & CSI                          & DNN                               & Y                \\ \cline{2-7} 
                                        & \cite{RN124}                   & D                & Infra Red Laser      & Video                        & -                                 & -                \\ \hline

\end{tabular}
\end{adjustbox}
\end{table*}

\end{normalsize}

\begin{normalsize}
\begin{table*}[!t]
\centering
\caption{Comparison of different approaches for recognizing motion based activities.\newline Symbols used: D= device-free, W= wearable, H= hybrid, Y= yes, N= no, -- = Not Available}
\label{my-label}
\begin{adjustbox}{width= \textwidth}
\begin{tabular}{|p{3cm} |p{1cm}|p{3cm} |p{1cm}|p{1cm}|p{1cm}|p{1cm}|}  

\hline

\multicolumn{7}{|l|}{M6 = Application, M7 = Cost, M8 = Accuracy, M9 = Latency, M10 = Real Time}                                                                      \\ \hline
\textbf{Category}                 & \textbf{Paper} & \textbf{M6}                                          & \textbf{M7} & \textbf{M8} & \textbf{M9}   & \textbf{M10} \\ \hline
\multirow{5}{*}{Tracking}         & \cite{RN7}             & Tracking                                             & Medium      & High        & $\sim$20 s    & Y            \\ \cline{2-7} 
                                  & \cite{RN11}             & Tracking                                             & Medium      & High        & -             & Y            \\ \cline{2-7} 
                                  & \cite{RN5}             & Tracking                                             & Low         & High        & -             & Y            \\ \cline{2-7} 
                                  & \cite{RN103}             & Tracking                                             & Low         & High        & -             & -            \\ \cline{2-7} 
                                  & \cite{RN104}             & Tracking                                             & Low         & High        & -             & -            \\ \hline
\multirow{5}{*}{Motion Detection} & \cite{RN111}             & Setting, walking, Fall                               & High        & High        & -             & -            \\ \cline{2-7} 
                                  & \cite{RN113}             & Motion Detection, Tracking                           & Low         & High        & -             & Y            \\ \cline{2-7} 
                                  & \cite{RN109}             & Human Presence Detection                             & Low         & High        & -             & -            \\ \cline{2-7} 
                                  & \cite{RN112}             & Motion Detection, Daily Activities (Posture Related) & High        & High        & 4.42 s        & -            \\ \cline{2-7} 
                                  & \cite{RN115}             & Detection of Stationary \& Moving Person, Direction  & Low         & High        & \textless 1 s & Y            \\ \hline
\multirow{4}{*}{People Counting}  & \cite{RN126}             & Counting People                                      & Low         & High        & -             & -            \\ \cline{2-7} 
                                  & \cite{RN123}             & Counting People                                      & High        & High        & -             & -            \\ \cline{2-7} 
                                  & \cite{RN125}             & Counting People                                      & Low         & Medium      & -             & Y            \\ \cline{2-7} 
                                  & \cite{RN124}              & Counting People, Tracking                            & High        & High        & -             & Y            \\ \hline

\end{tabular}
\end{adjustbox}
\end{table*}
\end{normalsize}

\subsubsection*{\textbf{Summary}}
A summary of the work presented for motion based activities is given in Table 13 and 14. We have tried to include device-free solutions. During the literature review, we found that RFID technology is leading the area of tracking and indoor localization. Most of the solutions presented for tracking and localization are using RFID tags deployed in the environment and a fixed reader with antenna is used to collect data from these tags. Most of these solutions are low cost and have high accuracy.

Besides RFID, solutions for motion detection and people counting are also using sensor-based and RF-based approaches. Different sensors like infrared and pressure sensors are used to detect the presence of people in a specific place. Radio Frequency technology is also used for research in motion detection and people counting. Use of Wi-Fi is one such example.

In our opinion, people counting is the sub-area of motion-based HAR in which the least amount of work is done. Tracking and localization are two such areas in which a significant amount of research has been done. One possible reason for less work in people counting would be the development of such sensors which can detect motion and can estimate the count with minimum processing required.

\subsection{Interaction Based Activities}
\label{interaction-based}
Some activities can also be performed by interacting with objects or using objects. A human can interact with objects in different ways. Interacting with objects in different ways results in different activities. Recognition of these activities is important in many applications (e.g., entertainment). In this section, we discuss some of the activities which are based on human object interaction.

\subsubsection{Human Object Interaction}
Human-computer interaction (HCI) is a flourishing area of research about the interaction between users and different machines. A considerable amount of work has been done in this field and completely new ways of interacting with machines have been proposed. The traditional method of interacting with a machine (e.g., computer), was through a mouse or keyboard but now there are many different ways of interacting with machines \cite{xie2018multi}. You can interact with a machine by making gestures or performing a specific activity. Users can perform these gestures or activity either by their own body or by using some objects i.e., by interacting with an object in a certain way, you can control a machine or provide input. Recently, different techniques have been proposed for interaction with machines, which are based on the interaction with objects.

RFID technology is playing an important role in the field of HCI. Main reasons for the use of RFID are: RFID passive tags are battery-free and do not need any maintenance, RFID tags are cheap as compared to other wireless sensors, and these tags can be easily attached to any object. Many solutions have been presented using RFID tags attached to objects for interacting with machines. RFID Shakable \cite{RN116} is a technique in which passive RFID tags are attached to different toys. The basic idea of this work is the pairing of two objects on the bases of their gestures. When two objects are tagged with RFID and they move in the vicinity of an RFID reader, information about their movement can be captured by the reader from those attached tags.  After applying gesture recognition, similar objects can be identified for pairing.

Li et al. \cite{RN117} proposed a technique called IDSense for detecting human-object interactions. The basic idea of IDSense is that it uses the changes in the signal parameters from RFID tags such as RSSI, Phase and Read Rate to detect human-object interaction. A single tag is attached to different objects and a reader antenna is used to investigate these tags when interacted by a human. Using SVM, IDSense can classify these interactions into different states of the object such as touch, still, swipe, and motion. Authors have demonstrated the application of this technique in three case studies which are: interactive storytelling with toys, interaction detection of the daily object for activity inferencing, and product interaction tracking in a superstore. The proposed technique is simple to implement and can provide results in real time but there are some limitations. Due to a single tag per object, similar interactions cannot be recognized correctly such as translation and rotation. This can be overcome by using multiple tags per object. The performance is also sensitive to the speed of the interaction i.e., too slow or too fast interactions may not be detected correctly.

Li et al.  \cite{RN118} presented a technique called PaperID through which a simple paper can be converted into an interactive input device using passive RFID tags. Different gestures like touch, swipe, cover, wave, slide and free air motion can be identified using this technique. The dense placing of multiple tags (on paper in this case) can cause interference in their signals. To overcome this problem, this work proposes a concept of half antenna in which the antenna of the tag is monopole i.e., only half of the antenna is present. As a result, the tag cannot harvest energy from the reader and is not readable. But when the antenna is completed (e.g., by touching), the tag becomes readable. This work also proposes techniques for making custom tags using conductive ink. Using these techniques, custom tags can be created very cheaply and on the spot, according to the need.

A similar technique called Rio is presented by Pradhan et al.\cite{RN119} through which any surface can be converted into a touchpad by attaching passive RFID tags. The basic theme of this technique is based on the change in impedance in tag antenna which occurs as a result of touching RFID tag. This change in impedance causes a phase change in the backscattered signal. Using this change in phase and machine learning algorithm, different gestures can be identified e.g., touch and swipe. The solution can work for both COTS and specially designed tags. No modification is required in the hardware. The technique can work for both single and multiple tags.

\begin{normalsize}
\begin{table*}[!t]
\centering
\caption{Comparison of different approaches for recognizing interaction based activities.\newline Symbols used: D= device-free, W= wearable, H= hybrid, O = object tagged, Y= yes, N= no, -- = Not Available}
\label{my-label}
\begin{adjustbox}{width= \linewidth}
\normalsize
\begin{tabular}{|p{2cm} |p{1.5cm}|p{1cm} |p{2.5cm}|p{4cm}|p{3.5cm}|p{0.5cm}|}  
\hline

\multicolumn{7}{|l|}{M1 = Approach ,M2 = Technology, M3 = Information Type, M4 = ML Algorithm , M5 = Supervised/Unsupervised}                         \\ \hline
\textbf{Category}                         & \textbf{Paper} & \textbf{M1} & \textbf{M2} & \textbf{M3}           & \textbf{M4}       & \textbf{M5} \\ \hline
\multirow{5}{*}{\begin{minipage}{2 cm}Human Object Interaction\end{minipage}} & \cite{RN116}              & O           & RFID        & Tag ID                & Cross-correlation & -           \\ \cline{2-7} 
                                          & \cite{RN117}             & O           & RFID        & RSSI, Phase, Tag ID   & SVM               & Y           \\ \cline{2-7} 
                                          & \cite{RN118}             & O           & RFID        & RSSI, Phase, ReadRate & SVM               & Y           \\ \cline{2-7} 
                                          & \cite{RN119}             & O           & RFID        & Phase Values          & -                 & Y           \\ \cline{2-7} 
                                          & \cite{RN128}             & O           & RFID        & Phase Values          & -                 & NA          \\ \hline

\end{tabular}
\end{adjustbox}
\end{table*}
\end{normalsize}

\begin{normalsize}
\begin{table*}[!t]
\centering
\caption{Comparison of different approaches for recognizing interaction based activities.\newline Symbols used: D= device-free, W= wearable, H= hybrid, O = object tagged, Y= yes, N= no, -- = Not Available}
\label{my-label}
\begin{adjustbox}{width= \textwidth}
\begin{tabular}{|p{2cm} |p{1cm}|p{4cm} |p{1cm}|p{1cm}|p{1cm}|p{1cm}|}  

\hline

\multicolumn{7}{|l|}{M6 = Application, M7 = Cost, M8 = Accuracy, M9 = Latency, M10 = RealTime}                                                                  \\ \hline
\textbf{Category}                         & \textbf{Paper} & \textbf{M6}                             & \textbf{M7} & \textbf{M8} & \textbf{M9}   & \textbf{M10} \\ \hline
\multirow{5}{*}{\begin{minipage}{2 cm}Human Object Interaction\end{minipage}} & \cite{RN116}             & Pairng                                  & Low         & High        & -             & -            \\ \cline{2-7} 
                                          & \cite{RN117}             & 4+ Types of Interactions                & Low         & High        & 1 s           & Y            \\ \cline{2-7} 
                                          & \cite{RN118}             & 5+ Types of Interactions                & Low         & High        & 0.5 s         & Y            \\ \cline{2-7} 
                                          & \cite{RN119}             & Touch, Track                            & Low         & High        & \textless 1 s & Y            \\ \cline{2-7} 
                                          & \cite{RN128}             & English Letters Recognition, 4 Gestures & Low         & High        & -             & Y            \\ \hline

\end{tabular}
\end{adjustbox}
\end{table*}
\end{normalsize}

Shangguan et al. \cite{RN71} presented the design and implementation of a technique called Pantomime which is capable of gesture recognition with only one antenna per location. This technique uses passive RFID tags attached to objects (two per object). When this object is moved in the air, the system is capable of recognizing the trajectory of the attached tag and thus the gesture made by the object can be recognized. By attaching two tags per object, the tag population becomes double causing a decrease in the reading rate of the tags. This will lead to a sparsely collected measurement which can affect the gesture recognition significantly. Also, a small gap between tags can cause the coupling effect which may lead to errors in phase values. Pantomime address these challenges by using a technique to boost the reading rate. In this technique, only the target tags are read and not the remaining tags in the coverage area. Pantomime identifies the target tag in a way that phase of that tag will go through a significant change when picked by a user and the phases of the rest of the tags will be stationary because of no disturbance. Once the target tags are identified, only values from those tags will be read and the remaining tags will be blocked, for a specific time (till these tags become stationary again), by manipulating EPC-standard C1G2 low-level interrogation process. The application of Pantomime is demonstrated by two case studies: handwriting tracking on whiteboard and superstore item querying. In the later case, a user picks any tagged-object at random and makes different pre-defined gestures with it, in front of a reader antenna. The system can recognize the gestures made with the object.

\subsubsection*{\textbf{Summary}}
Table 15 and 16 summarize the work presented for interaction based activities. This area has gained much popularity in recent years because of its application in various fields such as gaming, entertainment, and human-computer interaction. Many different approaches have been used for human-object interactions. We have tried to focus on device-free approaches. However, there are many solutions which use wearable approach.

Use of RFID is very common in recognition of interaction based activities because of its passive nature. The passive RFID tag can be attached to any object and can provide information via wireless communication to the reader. Using RFID technology, many solutions have been presented for smart surface, touchpads and gesture recognition using object interaction. These solutions are low cost (because RFID technology is cheap) and provide high accuracy.

Previously, specially designed hardware surface using different capacitors were used as an input device or touchpads. But now, with the help of research in human object interaction area, any common surface (e.g., paper) can be converted into a touchpad or a smart surface. Research in human object interaction area is providing new and interesting ways for communicating with machines (instead of typical methods such as keyboard and mouse). Now, interaction with machines is possible by performing certain gestures and interacting with objects in a certain manner.  

\section{Applications of Human Activity Recognition}

Human activity recognition is a very complex and challenging task. The basic goal of HAR is to interpret human activities by observing and analyzing the information collected about the activity. Interpreting and knowing human activities is of great importance and therefore, HAR has many applications in various fields \cite{RN148}. Following are some of the application areas of HAR as shown in Figure 8.

\begin{figure}[h]
	\centering
	\includegraphics[width=  \linewidth]{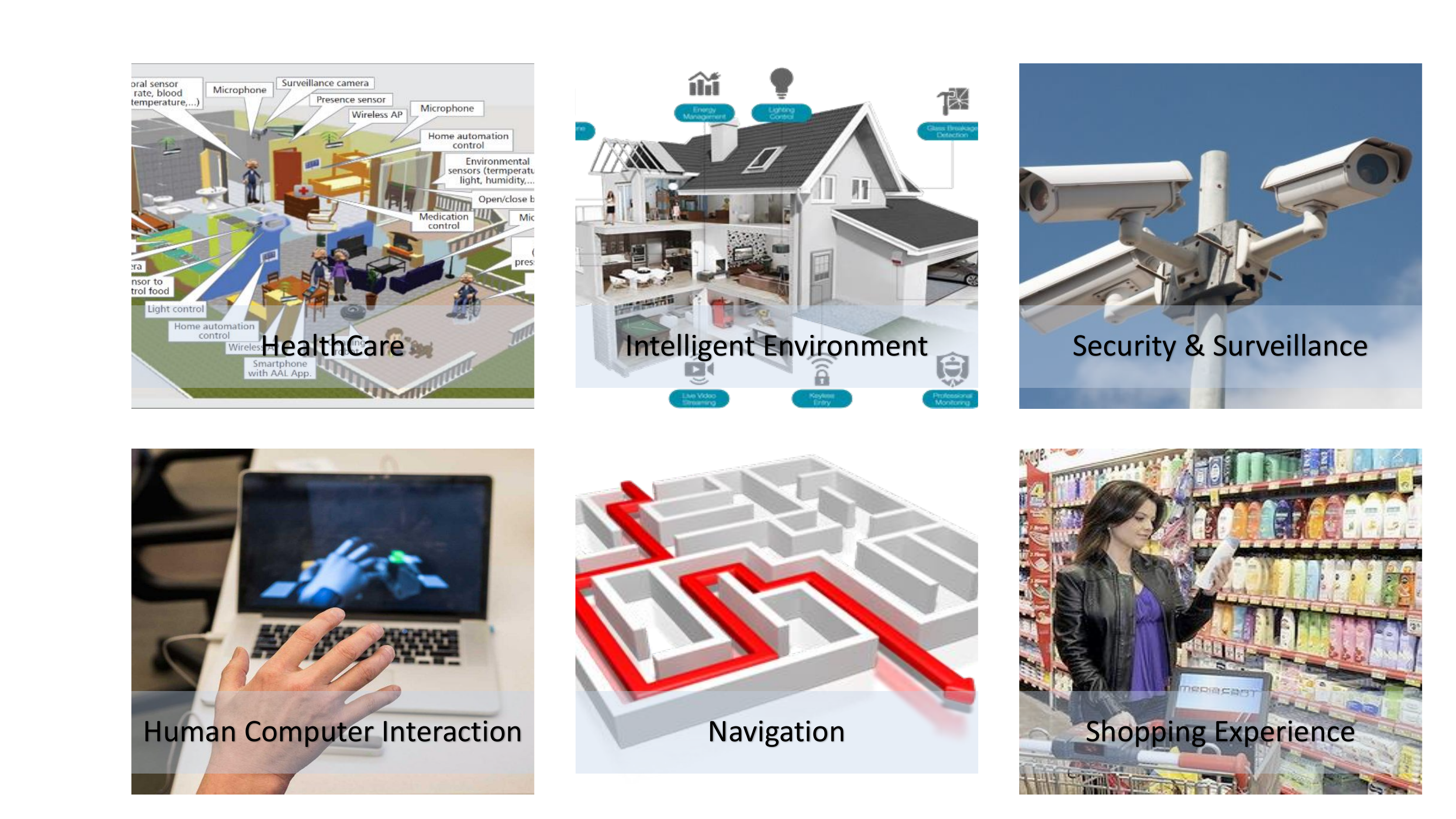}
	\caption{Some applications of human activity recognition}
	\label{fig1}
\end{figure}

\subsection{Elder Health Care }
With the advancement in medical science and technology, life expectancy has been increased. The population is aging around the world with an alarming rate. 15\% of the Australian population is 65 or over and this number will double by 2056 \cite{cdc}. Another report by Goldstone states that by the year 2050, 30\% of Chinese, European, Canadian and American people will have more than 60 years of age \cite{abc}. This increase in the number of elder people will cause many problems such as medical cost and shortage of caring staff. It will increase the demand for medical staff which can assist elder people in their daily life. To solve this issue, many solutions have been proposed in recent years. New technologies have been developed to help elder people live independently. These technologies are called Ambient Assisted Living (AAL) tools and are helping people with issues such as remote monitoring, fall detection, medication management, medication reminder, exercise management, and independent living \cite{RN45,RN49,RN54,RN37,RN139,RN140,RN141,RN142,RN145,adame2018cuidats,RN147}.  Human activity recognition (HAR) is an important part of AAL. With the help of recent advancement in the field of human activity recognition, it has become possible to monitor elder people remotely without having a human presence on the spot. Human activity recognition can help elder people to live independently. Especially, in case of patients like Alzheimer's and dementia, HAR can help in early detection and also assist doctors and caregivers in the treatment of these patients. By monitoring human activities and reporting any abnormal activity such as fall, HAR is helping to reduce medical expenses. HAR is also helping to reduce the demand for health givers which is not only costly but the world will face a shortage of trained professional because of the aging population.

\subsection{Intelligent Environment}

Intelligent environment refers to an environment in which the objects can sense and communicate. These environments have embedded systems and sensors which can capture and communicate information. Building smart environments is a very hot research topic nowadays. Human activity recognition is an essential part of smart environments such as smart homes, smart offices, smart health centers, and smart old care centers \cite{RN149,RN150,RN152}. A smart environment adapts itself according to the activities of its resident (e.g., turning off the lights when the residents are asleep). Therefore, it is important to know about the activities of residents and that is where human activity recognition comes in play. Through techniques of HAR, smart environments can know about the activity of its residents and can adapt itself accordingly. For example, if no one is present in a smart home, it will turn off the lights, cooling/heating system and will close the windows, etc. If the residents enter the home, the smart home will automatically detect their presence and will turn on the lights and other systems accordingly. In a smart health care center which is context-aware, activities of patients can be monitored remotely. Doctors can monitor the status of patients and can check their routines for exercise or any therapy. Smart care centers can support the independent living of elder people. As most of the elder people have different diseases and they have problems with mobility, a smart care center can assist in their independent living. 

\subsection{Security and Surveillance}
The traditional way of security and surveillance is to have a watchman guarding a facility. This approach is not practical for many reasons. Humans have limitations and needs and can make mistakes. The field of security and surveillance went through a paradigm shift and the role of security guards was taken by surveillance cameras. Unlike a security guard (human), surveillance cameras can watch 24x7 without any interruptions. However, there is still a need for human effort to monitor these security cameras for any suspicious activity. This limitation is overcome by research on human activity recognition. Different techniques have been proposed for automatic human activity recognition. These techniques are for both vision-based approaches and sensor-based approaches. For vision (cameras) based techniques, many solutions have been proposed which can analyze the video or image from the camera and can recognize the activities going on, thus, can report any suspicious activity without involving any human effort \cite{RN153}. Also, many solutions have been proposed which can analyze and process the information from different sensors deployed in a facility and can report any suspicious activity such as intrusion \cite{RN104,RN113}. 

\subsection{Human Computer Interaction}
Human activity recognition is playing a central role in the field of HCI. The traditional way of interacting with a computer or any machine is to use an input device like a keyboard and mouse. This approach is not practical in many situations. For example, in public places like hospitals, train stations, airport, and parks, it is not feasible to place input devices. Also, some people may have a problem with using these input devices e.g., elder people or patients. Recently, through research in human activity recognition, many techniques have been proposed for interacting with machines without using any input devices. Now you can interact with a machine by making free gestures in the air or perform a specific activity for giving a particular command to the machine. Machines are able to recognize and understand these commands. Many gesture recognition solutions have been proposed in the past decades. Some of these solutions use wearable approach \cite{RN67,RN66,xie2018multi} while some are device-free \cite{RN75,RN76,RN77}. These techniques have revolutionized the entertainment and gaming industry.  Thanks to HAR, gamers can interact with games by performing actual activities and the game is able to recognize these activities. HAR is also helping robots to interact with humans by recognizing their activities.  

\subsection{Indoor Navigation}

Global Positioning System-based outdoor navigation is a well-established area and has become a crucial part of our lives but indoor navigation is still an open issue. People spend most of their time inside buildings such as homes, offices, shopping malls, and restaurants. GPS doesn't work in indoor environments because of the low signal received inside the buildings (obstacles such as concrete, block the GPS signal). There should be an alternate solution for indoor navigation instead of using GPS like outdoor navigation. Navigation is a very important topic and has many applications which include localization and tracking, assisting elder people and people with disabilities, helping shoppers in a large shopping mall, assisting people in a big facility such as airports or hospitals, helping emergency response teams, etc. Research in human activity recognition is helping to solve the problem of indoor navigation. Many solutions have been proposed to detect human motion and presence, localize and track their movement and find the frequent trajectories in a certain environment \cite{RN101,RN5,RN99,RN113}. All these solutions come under the umbrella of human activity recognition research and are helping society in terms of navigation. 

\subsection{Shopping Experience}

Analyzing and understanding the shopping behavior of the customers is of great importance to business owners. Online stores can analyze customer's behavior very easily from the clicks and shopping carts of the customers. But analyzing the behavior of customers in a physical store is very challenging. The only data that business owners have about the shoppers, is sale history. Sale history can only provide details about the total profit or loss and the best seller item of the store. Business owners need information about the experience of shoppers.  This information includes items in which customer showed interest, items which customers ignored, items on which customer spent most of the time, items which customers browsed or bought together, items which customers compared before buying, brands which customers were interested more, etc. This kind of information can help the business owner to boost their business and also provide the best shopping experience to the customers.  Research in the area of human activity recognition has provided different solutions for recognizing the behaviors of shoppers. These solutions use different technologies. Some of these use surveillance cameras \cite{RN92} while others use Wi-Fi \cite{RN94}. With recent advancement in RFID technology, many solutions have been proposed for shopper's behavior analysis using RFID technology \cite{RN96,RN97}. In this way, HAR is helping business owners to enhance their business and provide a better shopping experience to the customers.

\section{Open Issues and Future Research Directions}

Although significant research has been done in the field of human activity recognition, there are some open issues which still need to be addressed. In this section, we present some of the open issues in human activity recognition.

\subsection{Complex Activities}
Existing work can recognize basic and atomic activities which are performed by a single subject. Also, the model needs to be trained for similar activities in advance. But there are many complex activities which existing solutions cannot recognize. Following are some types of activities which offer further research opportunities for researchers.  
\subsubsection*{Composite Activity} Most of the current solutions are focused on the recognition of simple activities performed by a single subject such as walking, running, eating, and sitting. But daily life is not only about these simple activities. There are many activities which are composite and consist of multiple simple activities. For example, doing exercise is a composite activity which consists of atomic activities such as sitting, standing, and running. Recognizing such a composite activity is very challenging as compared to the recognition of atomic activities. Blanke \& Schiele \cite{blanke2010remember} has discussed this issue in detail and has provided a potential solution for recognition of composite activities. 

\subsubsection*{Multiple Subjects}  Almost all solutions presented till now are capable of recognizing the activities of a single subject. Whether it is tracking, gesture recognition, posture recognition, or other areas, existing work is focused on recognizing the activity of a single person at a time. But in the real world, there are many situations in which activities are performed by multiple subjects simultaneously (e.g., people in the kitchen or living room) or multiple people involved in a single activity (handshake, hugging, etc.). Some researchers tried to solve this problem of recognizing activities in a multi-subject environment. Wang et al. \cite{wang2009sensor} proposed a solution for recognizing multi-users activities in a smart home, using dense sensing approach. Another work presented by Singla et al. \cite{singla2010recognizing}, proposed HMM model for recognizing the activities of two residents. But still, this problem is not solved completely and requires further research. 

\subsubsection*{Concurrent Activities}
Existing work is based on the assumption that a person will perform only one activity at a time. It can be true for ambulatory activities such as running and walking. But there are many situations in which users are performing concurrent activities i.e., multiple activities at the same time. For example, a person can be reading a newspaper while drinking coffee or having lunch while watching TV. Very little research has been done in this area and there is a potential for further work in this area. Further details about this challenge can be found in \cite{helaoui2011recognizing}. 

\subsubsection*{Variability}
Present solutions for activity recognition face the issue of variability. Variability means that if the same activity is performed by a different person or the same activity is performed by the same person at a different pace. Many existing systems cannot deal with variability problem i.e., if the same activity is performed by a different person, the system's recognition accuracy is very low. Also, if the same person performs the same activity in a different style, the system's performance degrades. Modern systems should be robust and should deal with the issue of variability. It is still an open issue and needs further research.

\subsection{Need for Intelligent Solutions}
Current solutions follow a traditional approach in which a model is trained for some activities and it can then recognize only that type of activities. Also, current HAR solutions are capable of recognizing only the past activities. But in today's world, there is a need for smart solutions which can detect normal activities (for which the model is trained) as well as abnormal activities and are also capable of predicting the future activities. Following are the two areas which need further research for making HAR solutions intelligent.

\subsubsection*{Abnormal Activities}
Existing solutions are focused on recognizing activities which are normal daily life activities such as sitting, standing, sleeping, walking, and eating. Recognition of these activities is important but that is not the whole purpose of activity recognition, especially for those applications which intend to identify abnormal activities. Detection of abnormal activities is of great importance in applications like security and healthcare. In security and surveillance, any abnormal activity is suspicious and should be reported immediately so that proper action can be taken. In healthcare, the detection of abnormal activity is very important for remote monitoring. If anything abnormal is detected, proper assistance should be provided. Recognition of abnormal activities is a challenging task due to many reasons. There is no single definition of abnormal activity and many interpretations are available to define abnormal activity. According to \cite{yin2008sensor}, abnormal activities occur rarely and are not expected in advance. Another hurdle in recognition of abnormal activities is the availability of data. For normal activities, a substantial amount of data is available to train the model but data for abnormal activity is very scared. There is a need for further research in this area. 

\subsubsection*{Predicting Next Activity}
Almost all of the existing solutions can recognize the past activity i.e., when activity happens, the given system can recognize it. This means that current HAR system can recognize previous activities, which is helpful in many situations. But an interesting thing would be that if the HAR system can predict future activity i.e., what will happen next. This function is very important, especially in applications like fall detection/prevention. If a HAR system can tell the caregivers that a patient or an elder person is about to fall, fall can be prevented, which is very helpful. A possible research direction would be to make the HAR system not only recognize the current and past activities but should also predict future activities.

\subsection{Environmental Interference and Experimental Setup}
Significant research work has been conducted for HAR but still dealing with environmental interference is a big challenge.  Also, the existing solutions are labor intensive and need extensive training before testing. Currently, there is no benchmark (in terms of data and experimental setup) for evaluating the performance of HAR techniques. These areas offer future research opportunities and are discussed in this section.

\subsubsection*{Requirement of Extensive Training}
Almost all the solutions proposed, required training. Getting training data is not an easy job, especially in the case of elder people. Many of these solutions are heavily dependent on training and required to be trained again if the environment is changed. For example, if you need to implement it in a different room or home, you have to train it again. Also, the training time for some solutions is too long and need to be trained offline. A good solution for activity recognition should be independent of the environment i.e., once trained, it should work in any similar environment. This aspect of HAR system needs further research.

\subsubsection*{Environmental Interference}
Although the research in human activity recognition using device-free approach has become very advanced in many ways, still, dealing with environmental interference is an issue. Most of the solutions proposed are vulnerable to environmental factors and their performance is affected by the outside world. The device-free approach is getting more attention because of its advantage as users are not required to carry any device with them but dealing with environmental interference is still an open issue and requires more research to minimize its influence. 

\subsubsection*{Need for Standard Testing Setup}
Solutions proposed for different sub-areas of activity recognition use different approaches. The experimental setup is different and the environment is different. Therefore, it is very difficult to compare these techniques for evaluation. There is no standard set up or benchmark (e.g., benchmark data sets in data mining) for evaluating the performance of a solution. There is a need for such a system, through which the performance of any new or existing solution can be evaluated.

\subsection{Security of the System}
In the literature review of the human activity recognition system, almost all the solutions have ignored the security aspect. The proposed solution are more focused on accuracy, cost, scalability, etc., while ignoring the aspect of security. Security is an important aspect of the human activity recognition system. Information about the activity of a person should be available to authorized people only. Discussion about accessibility, privacy, and security of the information about human activities is missing from the literature and this area needs to be investigated.

\section{Discussion}

In this article, we discussed and analyzed different aspects of human activity recognition. We presented a review of the overall work conducted in different areas of the activity recognition with main focus on device-free approaches. As obvious from section 4, different approaches have been used for recognizing the activities of human. We found that the comparison of these techniques is difficult due to the following reasons. 

\subsubsection*{\textbf{Main Focus}}We found that comparing these techniques is difficult due to various reasons. As shown in Figure 6, we have divided human activity recognition research into different sub-areas. All these sub-areas come under the umbrella of activity recognition. We have provided a literature review for all these sub-areas and have covered different techniques proposed in these sub-areas. The main focus of the work discussed varies, as some of them focus on one sub-area while others focus on another sub-area. Because of this, it is difficult to compare all these techniques. For example, comparing a technique for gesture recognition with a technique for ADL recognition is difficult. In gesture recognition, the processing time is very important and the solution needs to provide the results in real time while in the case of ADL, time is not a big issue instead importance is given to the accuracy of the results. However, we have tried to provide the readers, a comparison of these techniques on some common ground. 

\subsubsection*{\textbf{Approach}}Different techniques use different hardware. For example, some techniques use wearable devices while others use device-free approach, some use sensors attached to the objects while others use Wi-Fi. Also, these solutions use different classification methods (machine learning tools). Comparing such solutions, which are based on completely different approaches is not an easy job.  Every approach has its pros and cons but comparing these approaches with others, is challenging. We tried our best to provide a detailed comparison to the reader.

\subsubsection*{\textbf{Experimental Setup}}There is no universal setup for evaluating these techniques. Experimental setups used in different solutions are different from each other. For example, some solutions use wearable devices and perform experiments in a room while other approaches use tagged-objects and perform experiments in a kitchen. Comparing solutions with the different experimental setup is challenging because accuracy and other factors depend on the experimental environment.

\subsubsection*{\textbf{Missing Details}}A major issue that we faced in our literature review, is the missing details, as you can see from Table 2, 3 and 4. There are some papers which lack information about very important things. For example, most of the papers lack the discussion about time and space complexity of their techniques. There is no discussion about the latency of the proposed approach which is a very important factor in activity recognition. Some papers are missing the details about the classifier (machine learning algorithm) used in the approach.  There are papers in which there is confusion about the working of the proposed technique such as, whether the proposed technique is real-time or not and off-line or online. Authors should try to provide detailed information about everything involved in their approach. They should include a discussion section to provide details such as latency, complexity, and limitation. 

\section{Conclusion}
In this work, we have presented a comprehensive overview of the research work in human activity recognition. Unlike other surveys which focus only on a single type of activities, we covered almost all the sub-fields of activity recognition. We divided the research in activity recognition into three main categories: action-based, motion-based and interaction-based. We further divided these into 10 different sub-categories and presented the latest literature for each category. The Main focus of this survey is device-free approaches with a focus on RFID technology. We discussed the latest literature using device-free approach for human activity recognition and provided a comprehensive comparison of the different techniques included in the literature review. We also provided some applications of human activity recognition in various areas. In the end, we discussed some open research issues in activity recognition and provided future research directions.

\bibliographystyle{IEEEtran}
\bibliography{SurveyPaper}

\begin{thebibliography}{100}
\providecommand{\url}[1]{#1}
\csname url@rmstyle\endcsname
\providecommand{\newblock}{\relax}
\providecommand{\bibinfo}[2]{#2}
\providecommand\BIBentrySTDinterwordspacing{\spaceskip=0pt\relax}
\providecommand\BIBentryALTinterwordstretchfactor{4}
\providecommand\BIBentryALTinterwordspacing{\spaceskip=\fontdimen2\font plus
\BIBentryALTinterwordstretchfactor\fontdimen3\font minus
  \fontdimen4\font\relax}
\providecommand\BIBforeignlanguage[2]{{%
\expandafter\ifx\csname l@#1\endcsname\relax
\typeout{** WARNING: IEEEtran.bst: No hyphenation pattern has been}%
\typeout{** loaded for the language `#1'. Using the pattern for}%
\typeout{** the default language instead.}%
\else
\language=\csname l@#1\endcsname
\fi
#2}}

\bibitem{yang2011activity}
J.~Yang, J.~Lee, and J.~Choi, ``Activity recognition based on rfid object usage
  for smart mobile devices,'' \emph{Journal of Computer Science and
  Technology}, vol.~26, no.~2, pp. 239--246, 2011.

\bibitem{RN79}
L.~Chen, J.~Hoey, C.~D. Nugent, D.~J. Cook, and Z.~Yu, ``Sensor-based activity
  recognition,'' \emph{IEEE Transactions on Systems, Man, and Cybernetics, Part
  C (Applications and Reviews)}, vol.~42, no.~6, pp. 790--808, 2012.

\bibitem{aggarwal2014human}
J.~K. Aggarwal and L.~Xia, ``Human activity recognition from 3d data: A
  review,'' \emph{Pattern Recognition Letters}, vol.~48, pp. 70--80, 2014.

\bibitem{RN83}
S.~Wang and G.~Zhou, ``A review on radio based activity recognition,''
  \emph{Digital Communications and Networks}, vol.~1, no.~1, pp. 20--29, 2015.

\bibitem{RN78}
M.~Scholz, S.~Sigg, H.~R. Schmidtke, and M.~Beigl, ``Challenges for device-free
  radio-based activity recognition,'' in \emph{Workshop on Context Systems,
  Design, Evaluation and Optimisation}, 2011, Conference Proceedings.

\bibitem{RN81}
S.~Amendola, R.~Lodato, S.~Manzari, C.~Occhiuzzi, and G.~Marrocco, ``Rfid
  technology for iot-based personal healthcare in smart spaces,'' \emph{IEEE
  Internet of things journal}, vol.~1, no.~2, pp. 144--152, 2014.

\bibitem{RN86}
J.~Ma, H.~Wang, D.~Zhang, Y.~Wang, and Y.~Wang, ``A survey on wi-fi based
  contactless activity recognition,'' in \emph{Intl IEEE Conferences on
  Ubiquitous Intelligence \& Computing, Advanced and Trusted Computing,
  Scalable Computing and Communications, Cloud and Big Data Computing, Internet
  of People, and Smart World Congress
  (UIC/ATC/ScalCom/CBDCom/IoP/SmartWorld)}.\hskip 1em plus 0.5em minus
  0.4em\relax IEEE, 2016, pp. 1086--1091.

\bibitem{RN87}
E.~Cianca, M.~De~Sanctis, and S.~Di~Domenico, ``Radios as sensors,'' \emph{IEEE
  Internet of Things Journal}, vol.~4, no.~2, pp. 363--373, 2017.

\bibitem{RN90}
J.~Wang, Y.~Chen, S.~Hao, X.~Peng, and L.~Hu, ``Deep learning for sensor-based
  activity recognition: A survey,'' \emph{arXiv preprint arXiv:1707.03502},
  2017.

\bibitem{RN80}
O.~D. Lara and M.~A. Labrador, ``A survey on human activity recognition using
  wearable sensors,'' \emph{IEEE Communications Surveys and Tutorials},
  vol.~15, no.~3, pp. 1192--1209, 2013.

\bibitem{RN89}
M.~Cornacchia, K.~Ozcan, Y.~Zheng, and S.~Velipasalar, ``A survey on activity
  detection and classification using wearable sensors,'' \emph{IEEE Sensors
  Journal}, vol.~17, no.~2, pp. 386--403, 2017.

\bibitem{RN85}
M.~Shoaib, S.~Bosch, O.~D. Incel, H.~Scholten, and P.~J. Havinga, ``A survey of
  online activity recognition using mobile phones,'' \emph{Sensors}, vol.~15,
  no.~1, pp. 2059--2085, 2015.

\bibitem{RN84}
M.~Vrigkas, C.~Nikou, and I.~A. Kakadiaris, ``A review of human activity
  recognition methods,'' \emph{Frontiers in Robotics and AI}, vol.~2, p.~28,
  2015.

\bibitem{RN88}
S.~Herath, M.~Harandi, and F.~Porikli, ``Going deeper into action recognition:
  A survey,'' \emph{Image and vision computing}, vol.~60, pp. 4--21, 2017.

\bibitem{maret2018real}
Y.~Maret, D.~Oberson, and M.~Gavrilova, ``Real-time embedded system for gesture
  recognition,'' in \emph{IEEE International Conference on Systems, Man, and
  Cybernetics (SMC)}.\hskip 1em plus 0.5em minus 0.4em\relax IEEE, 2018, pp.
  30--34.

\bibitem{landt2005history}
J.~Landt, ``The history of rfid,'' \emph{IEEE potentials}, vol.~24, no.~4, pp.
  8--11, 2005.

\bibitem{wu2011rfid}
Y.~Wu, D.~C. Ranasinghe, Q.~Z. Sheng, S.~Zeadally, and J.~Yu, ``Rfid enabled
  traceability networks: a survey,'' \emph{Distributed and Parallel Databases},
  vol.~29, no. 5-6, pp. 397--443, 2011.

\bibitem{ko2017accessibility}
C.-H. Ko, ``Accessibility of radio frequency identification technology in
  facilities maintenance.'' \emph{Journal of Engineering, Project \& Production
  Management}, vol.~7, no.~1, 2017.

\bibitem{RN57}
M.~Daniels, K.~Muldawer, J.~Schlessman, B.~Ozer, and W.~Wolf, ``Real-time human
  motion detection with distributed smart cameras,'' in \emph{First ACM/IEEE
  International Conference on Distributed Smart Cameras (ICDSC'07).}\hskip 1em
  plus 0.5em minus 0.4em\relax IEEE, Conference Proceedings, pp. 187--194.

\bibitem{RN58}
P.~Garg, N.~Aggarwal, and S.~Sofat, ``Vision based hand gesture recognition,''
  \emph{World Academy of Science, Engineering and Technology}, vol.~49, no.~1,
  pp. 972--977, 2009.

\bibitem{RN59}
Y.~Yao and Y.~Fu, ``Contour model-based hand-gesture recognition using the
  kinect sensor,'' \emph{IEEE transactions on circuits and systems for video
  technology}, vol.~24, no.~11, pp. 1935--1944, 2014.

\bibitem{RN60}
E.~Ohn-Bar and M.~M. Trivedi, ``Hand gesture recognition in real time for
  automotive interfaces: A multimodal vision-based approach and evaluations,''
  \emph{IEEE transactions on intelligent transportation systems}, vol.~15,
  no.~6, pp. 2368--2377, 2014.

\bibitem{RN61}
M.~H. Ali, F.~Kamaruzaman, M.~A. Rahman, and A.~A. Shafie, ``Automated secure
  room system,'' in \emph{4th International Conference on Software Engineering
  and Computer Systems (ICSECS)}.\hskip 1em plus 0.5em minus 0.4em\relax IEEE,
  2015, Conference Proceedings, pp. 73--78.

\bibitem{RN62}
N.~Sreekanth and N.~Narayanan, ``Dynamic gesture recognition - a machine vision
  based approach,'' in \emph{International Conference on Signal, Networks,
  Computing, and Systems}.\hskip 1em plus 0.5em minus 0.4em\relax Springer,
  2017, Conference Proceedings, pp. 105--115.

\bibitem{RN63}
R.~R. Itkarkar, A.~Nandi, and B.~Mane, \emph{Contour-Based Real-Time Hand
  Gesture Recognition for Indian Sign Language}.\hskip 1em plus 0.5em minus
  0.4em\relax Springer, 2017, pp. 683--691.

\bibitem{RN64}
J.~Singha, A.~Roy, and R.~H. Laskar, ``Dynamic hand gesture recognition using
  vision-based approach for human–computer interaction,'' \emph{Neural
  Computing and Applications}, vol.~29, no.~4, pp. 1129--1141, 2018.

\bibitem{RN65}
J.~Wang, D.~Vasisht, and D.~Katabi, ``Rf-idraw: virtual touch screen in the air
  using rf signals,'' in \emph{ACM SIGCOMM Computer Communication Review},
  vol.~44.\hskip 1em plus 0.5em minus 0.4em\relax ACM, 2014, Conference
  Proceedings, pp. 235--246.

\bibitem{RN67}
N.~Siddiqui and R.~H. Chan, ``A wearable hand gesture recognition device based
  on acoustic measurements at wrist,'' in \emph{39th Annual International
  Conference of Engineering in Medicine and Biology Society (EMBC)}.\hskip 1em
  plus 0.5em minus 0.4em\relax IEEE, 2017, Conference Proceedings, pp.
  4443--4446.

\bibitem{RN68}
F.-T. Liu, Y.-T. Wang, and H.-P. Ma, ``Gesture recognition with wearable 9-axis
  sensors,'' in \emph{IEEE International Conference on Communications
  (ICC)}.\hskip 1em plus 0.5em minus 0.4em\relax IEEE, 2017, Conference
  Proceedings, pp. 1--6.

\bibitem{RN66}
L.~Xie, C.~Wang, A.~X. Liu, J.~Sun, and S.~Lu, ``Multi-touch in the air:
  Concurrent micromovement recognition using rf signals,'' \emph{IEEE/ACM
  Transactions on Networking (TON)}, vol.~26, no.~1, pp. 231--244, 2018.

\bibitem{RN69}
P.~Asadzadeh, L.~Kulik, and E.~Tanin, ``Gesture recognition using rfid
  technology,'' \emph{Personal and Ubiquitous Computing}, vol.~16, no.~3, pp.
  225--234, 2012.

\bibitem{RN70}
K.~Bouchard, A.~Bouzouane, and B.~Bouchard, ``Gesture recognition in smart home
  using passive rfid technology,'' in \emph{7th International Conference on
  Pervasive Technologies Related to Assistive Environments}.\hskip 1em plus
  0.5em minus 0.4em\relax ACM, 2014, Conference Proceedings, p.~12.

\bibitem{RN71}
L.~Shangguan, Z.~Zhou, and K.~Jamieson, ``Enabling gesture-based interactions
  with objects,'' in \emph{15th Annual International Conference on Mobile
  Systems, Applications, and Services}.\hskip 1em plus 0.5em minus 0.4em\relax
  ACM, 2017, Conference Proceedings, pp. 239--251.

\bibitem{RN72}
A.~Jayatilaka and D.~C. Ranasinghe, ``Real-time fluid intake gesture
  recognition based on batteryless uhf rfid technology,'' \emph{Pervasive and
  Mobile Computing}, vol.~34, pp. 146--156, 2017.

\bibitem{RN73}
L.-H. Chen, K.-C. Liu, C.-Y. Hsieh, and C.-T. Chan, ``Drinking gesture spotting
  and identification using single wrist-worn inertial sensor,'' in
  \emph{International Conference on Applied System Innovation (ICASI)}.\hskip
  1em plus 0.5em minus 0.4em\relax IEEE, 2017, Conference Proceedings, pp.
  299--302.

\bibitem{RN74}
S.~Ye, H.~Zeng, J.~Fan, and X.~Wang, ``Lsi-rec: A link state indicator based
  gesture recognition scheme in a rfid system,'' in \emph{9th Conference on
  Industrial Electronics and Applications (ICIEA)}.\hskip 1em plus 0.5em minus
  0.4em\relax IEEE, 2014, Conference Proceedings, pp. 406--411.

\bibitem{RN75}
R.~Parada, K.~Nur, J.~Melia-Segui, and R.~Pous, ``Smart surface: Rfid-based
  gesture recognition using k-means algorithm,'' in \emph{12th International
  Conference on Intelligent Environments (IE)}.\hskip 1em plus 0.5em minus
  0.4em\relax IEEE, 2016, Conference Proceedings, pp. 111--118.

\bibitem{RN76}
H.~Ding, C.~Qian, J.~Han, G.~Wang, W.~Xi, K.~Zhao, and J.~Zhao, ``Rfipad:
  Enabling cost-efficient and device-free in-air handwriting using passive
  tags,'' in \emph{IEEE 37th International Conference on Distributed Computing
  Systems (ICDCS)}.\hskip 1em plus 0.5em minus 0.4em\relax IEEE, 2017,
  Conference Proceedings, pp. 447--457.

\bibitem{RN77}
Y.~Zou, J.~Xiao, J.~Han, K.~Wu, Y.~Li, and L.~M. Ni, ``Grfid: A device-free
  rfid-based gesture recognition system,'' \emph{IEEE Transactions on Mobile
  Computing}, vol.~16, no.~2, pp. 381--393, 2017.

\bibitem{torres2015accelerometer}
C.~Torres-Huitzil and A.~Alvarez-Landero, ``Accelerometer-based human activity
  recognition in smartphones for healthcare services,'' in \emph{Mobile
  Health}.\hskip 1em plus 0.5em minus 0.4em\relax Springer, 2015, pp. 147--169.

\bibitem{RN22}
A.~Wickramasinghe and D.~C. Ranasinghe, ``Ambulatory monitoring using passive
  computational rfid sensors,'' \emph{IEEE Sensors Journal}, vol.~15, no.~10,
  pp. 5859--5869, 2015.

\bibitem{RN23}
C.~A. Ronao and S.-B. Cho, ``Human activity recognition with smartphone sensors
  using deep learning neural networks,'' \emph{Expert Systems with
  Applications}, vol.~59, pp. 235--244, 2016.

\bibitem{RN25}
D.~Castro, W.~Coral, C.~Rodriguez, J.~Cabra, and J.~Colorado, ``Wearable-based
  human activity recognition using and iot approach,'' \emph{Journal of Sensor
  and Actuator Networks}, vol.~6, no.~4, p.~28, 2017.

\bibitem{RN26}
A.~Ignatov, ``Real-time human activity recognition from accelerometer data
  using convolutional neural networks,'' \emph{Applied Soft Computing},
  vol.~62, pp. 915--922, 2018.

\bibitem{RN27}
B.~BENAISSA, M.~KÖPPEN, and K.~YOSHIDA, ``Activity and emotion recognition for
  elderly health monitoring,'' \emph{International Journal of Affective
  Engineering}, pp. IJAE--D--17--00\,020, 2017.

\bibitem{RN28}
F.~Xiao, Q.~Miao, X.~Xie, L.~Sun, and R.~Wang, ``Shmo: A seniors health
  monitoring system based on energy-free sensing,'' \emph{Computer Networks},
  2018.

\bibitem{RN13}
L.~Yao, Q.~Z. Sheng, W.~Ruan, T.~Gu, X.~Li, N.~Falkner, and Z.~Yang, ``Rf-care:
  Device-free posture recognition for elderly people using a passive rfid tag
  array,'' in \emph{12th EAI International Conference on Mobile and Ubiquitous
  Systems: Computing, Networking and Services}.\hskip 1em plus 0.5em minus
  0.4em\relax ICST (Institute for Computer Sciences, Social-Informatics and
  Telecommunications Engineering), 2015, Conference Proceedings, pp. 120--129.

\bibitem{RN19}
L.~Yao, Q.~Z. Sheng, X.~Li, T.~Gu, M.~Tan, X.~Wang, S.~Wang, and W.~Ruan,
  ``Compressive representation for device-free activity recognition with
  passive rfid signal strength,'' \emph{IEEE Transactions on Mobile Computing},
  vol.~17, no.~2, pp. 293--306, 2018.

\bibitem{li2018r}
L.~Li, R.~Bai, B.~Xie, Y.~Peng, A.~Wang, W.~Wang, B.~Jiang, J.~Liang, and
  X.~Chen, ``R\&p: An low-cost device-free activity recognition for e-health,''
  \emph{IEEE Access}, vol.~6, pp. 81--90, 2018.

\bibitem{RN17}
L.~Zhong, S.~Cho, D.~Metaxas, S.~Paris, and J.~Wang, ``Handling noise in single
  image deblurring using directional filters,'' in \emph{Conference on Computer
  Vision and Pattern Recognition (CVPR)}.\hskip 1em plus 0.5em minus
  0.4em\relax IEEE, 2013, Conference Proceedings, pp. 612--619.

\bibitem{RN18}
E.~J. Candes, C.~A. Sing-Long, and J.~D. Trzasko, ``Unbiased risk estimates for
  singular value thresholding and spectral estimators,'' \emph{IEEE
  transactions on signal processing}, vol.~61, no.~19, pp. 4643--4657, 2013.

\bibitem{avrahami2018below}
D.~Avrahami, M.~Patel, Y.~Yamaura, and S.~Kratz, ``Below the surface:
  Unobtrusive activity recognition for work surfaces using rf-radar sensing,''
  in \emph{23rd International Conference on Intelligent User Interfaces}.\hskip
  1em plus 0.5em minus 0.4em\relax ACM, 2018, pp. 439--451.

\bibitem{RN91}
S.-L. Chua, S.~Marsland, and H.~W. Guesgen, ``Behaviour recognition from
  sensory streams in smart environments,'' in \emph{Australasian Joint
  Conference on Artificial Intelligence}.\hskip 1em plus 0.5em minus
  0.4em\relax Springer, Conference Proceedings, pp. 666--675.

\bibitem{RN92}
M.~Popa, L.~Rothkrantz, Z.~Yang, P.~Wiggers, R.~Braspenning, and C.~Shan,
  ``Analysis of shopping behavior based on surveillance system,'' in \emph{IEEE
  International Conference on Systems Man and Cybernetics (SMC)}.\hskip 1em
  plus 0.5em minus 0.4em\relax IEEE, 2010, Conference Proceedings, pp.
  2512--2519.

\bibitem{RN93}
M.~Popa, A.~K. Koc, L.~J. Rothkrantz, C.~Shan, and P.~Wiggers, ``Kinect sensing
  of shopping related actions,'' in \emph{International Joint Conference on
  Ambient Intelligence}.\hskip 1em plus 0.5em minus 0.4em\relax Springer, 2011,
  Conference Proceedings, pp. 91--100.

\bibitem{RN94}
Y.~Zeng, P.~H. Pathak, and P.~Mohapatra, ``Analyzing shopper's behavior through
  wifi signals,'' in \emph{2nd workshop on Workshop on Physical
  Analytics}.\hskip 1em plus 0.5em minus 0.4em\relax ACM, 2015, Conference
  Proceedings, pp. 13--18.

\bibitem{RN96}
J.~Han, H.~Ding, C.~Qian, W.~Xi, Z.~Wang, Z.~Jiang, L.~Shangguan, and J.~Zhao,
  ``Cbid: A customer behavior identification system using passive tags,''
  \emph{IEEE/ACM Transactions on Networking}, vol.~24, no.~5, pp. 2885--2898,
  2016.

\bibitem{RN97}
Z.~Zhou, L.~Shangguan, X.~Zheng, L.~Yang, and Y.~Liu, ``Design and
  implementation of an rfid-based customer shopping behavior mining system,''
  \emph{IEEE/ACM transactions on networking}, vol.~25, no.~4, pp. 2405--2418,
  2017.

\bibitem{RN33}
N.~Noury, A.~Fleury, P.~Rumeau, A.~Bourke, G.~Laighin, V.~Rialle, and J.~Lundy,
  ``Fall detection-principles and methods,'' in \emph{29th Annual International
  Conference of Engineering in Medicine and Biology Society, (EMBS)}.\hskip 1em
  plus 0.5em minus 0.4em\relax IEEE, 2007, Conference Proceedings, pp.
  1663--1666.

\bibitem{RN35}
D.~Oliver, F.~Healey, and T.~P. Haines, ``Preventing falls and fall-related
  injuries in hospitals,'' \emph{Clinics in geriatric medicine}, vol.~26,
  no.~4, pp. 645--692, 2010.

\bibitem{RN36}
C.~S. Florence, G.~Bergen, A.~Atherly, E.~Burns, J.~Stevens, and C.~Drake,
  ``Medical costs of fatal and nonfatal falls in older adults,'' \emph{Journal
  of the American Geriatrics Society}, 2018.

\bibitem{RN34}
D.~Wild, U.~Nayak, and B.~Isaacs, ``How dangerous are falls in old people at
  home?'' \emph{Br Med J (Clin Res Ed)}, vol. 282, no. 6260, pp. 266--268,
  1981.

\bibitem{cdc}
{Center for Disease Control and Prevention}, ``{Web-based Injury Statistics
  Query and Reporting System},''
  \url{https://www.cdc.gov/homeandrecreationalsafety/falls/fallcost.html},
  2018, online; accessed 03 July 2018.

\bibitem{RN42}
S.-H. Cheng, ``An intelligent fall detection system using triaxial
  accelerometer integrated by active rfid,'' in \emph{International Conference
  on Machine Learning and Cybernetics (ICMLC)}, vol.~2.\hskip 1em plus 0.5em
  minus 0.4em\relax IEEE, 2014, Conference Proceedings, pp. 517--522.

\bibitem{RN43}
P.~Tsinganos and A.~Skodras, ``A smartphone-based fall detection system for the
  elderly,'' in \emph{10th International Symposium on Image and Signal
  Processing and Analysis (ISPA)}.\hskip 1em plus 0.5em minus 0.4em\relax IEEE,
  2017, Conference Proceedings, pp. 53--58.

\bibitem{RN44}
P.~Jatesiktat and W.~T. Ang, ``An elderly fall detection using a wrist-worn
  accelerometer and barometer,'' in \emph{39th Annual International Conference
  of Engineering in Medicine and Biology Society (EMBC)}.\hskip 1em plus 0.5em
  minus 0.4em\relax IEEE, 2017, Conference Proceedings, pp. 125--130.

\bibitem{RN45}
T.~N. Gia, V.~K. Sarker, I.~Tcarenko, A.~M. Rahmani, T.~Westerlund,
  P.~Liljeberg, and H.~Tenhunen, ``Energy efficient wearable sensor node for
  iot-based fall detection systems,'' \emph{Microprocessors and Microsystems},
  vol.~56, pp. 34--46, 2018.

\bibitem{RN46}
I.~Putra, J.~Brusey, E.~Gaura, and R.~Vesilo, ``An event-triggered machine
  learning approach for accelerometer-based fall detection,'' \emph{Sensors},
  vol.~18, no.~1, p.~20, 2017.

\bibitem{RN47}
G.~Rescio, A.~Leone, and P.~Siciliano, ``Supervised machine learning scheme for
  electromyography-based pre-fall detection system,'' \emph{Expert Systems with
  Applications}, vol. 100, pp. 95--105, 2018.

\bibitem{RN48}
D.~Zhang, H.~Wang, Y.~Wang, and J.~Ma, ``Anti-fall: A non-intrusive and
  real-time fall detector leveraging csi from commodity wifi devices,'' ser.
  Inclusive Smart Cities and e-Health.\hskip 1em plus 0.5em minus 0.4em\relax
  Springer International Publishing, 2015, Conference Proceedings, pp.
  181--193.

\bibitem{RN49}
H.~Wang, D.~Zhang, Y.~Wang, J.~Ma, Y.~Wang, and S.~Li, ``Rt-fall: A real-time
  and contactless fall detection system with commodity wifi devices,''
  \emph{IEEE Transactions on Mobile Computing}, vol.~16, no.~2, pp. 511--526,
  2017.

\bibitem{RN50}
Y.~Wang, K.~Wu, and L.~M. Ni, ``Wifall: Device-free fall detection by wireless
  networks,'' \emph{IEEE Transactions on Mobile Computing}, vol.~16, no.~2, pp.
  581--594, 2017.

\bibitem{RN51}
L.~Minvielle, M.~Atiq, R.~Serra, M.~Mougeot, and N.~Vayatis, ``Fall detection
  using smart floor sensor and supervised learning,'' in \emph{39th Annual
  International Conference of Engineering in Medicine and Biology Society
  (EMBC)}.\hskip 1em plus 0.5em minus 0.4em\relax IEEE, 2017, Conference
  Proceedings, pp. 3445--3448.

\bibitem{RN52}
S.~Kianoush, S.~Savazzi, F.~Vicentini, V.~Rampa, and M.~Giussani, ``Device-free
  rf human body fall detection and localization in industrial workplaces,''
  \emph{IEEE Internet of Things Journal}, vol.~4, no.~2, pp. 351--362, 2017.

\bibitem{RN54}
A.~Wickramasinghe, R.~L.~S. Torres, and D.~C. Ranasinghe, ``Recognition of
  falls using dense sensing in an ambient assisted living environment,''
  \emph{Pervasive and mobile computing}, vol.~34, pp. 14--24, 2017.

\bibitem{RN55}
R.~L.~S. Torres, A.~Wickramasinghe, V.~N. Pham, and D.~C. Ranasinghe, ``What if
  your floor could tell someone you fell? a device free fall detection
  method,'' in \emph{Conference on Artificial Intelligence in Medicine in
  Europe}.\hskip 1em plus 0.5em minus 0.4em\relax Springer, 2015, Conference
  Proceedings, pp. 86--95.

\bibitem{RN37}
W.~Ruan, L.~Yao, Q.~Z. Sheng, N.~Falkner, X.~Li, and T.~Gu, ``Tagfall: Towards
  unobstructive fine-grained fall detection based on uhf passive rfid tags,''
  in \emph{12th EAI International Conference on Mobile and Ubiquitous Systems:
  Computing, Networking and Services (MOBIQUITOUS)}.\hskip 1em plus 0.5em minus
  0.4em\relax ICST (Institute for Computer Sciences, Social-Informatics and
  Telecommunications Engineering), 2015, Conference Proceedings, pp. 140--149.

\bibitem{RN128}
X.~Xu, J.~Tang, X.~Zhang, X.~Liu, H.~Zhang, and Y.~Qiu, ``Exploring techniques
  for vision based human activity recognition: Methods, systems, and
  evaluation,'' \emph{Sensors}, vol.~13, no.~2, pp. 1635--1650, 2013.

\bibitem{RN129}
Y.~Yan, E.~Ricci, G.~Liu, and N.~Sebe, ``Egocentric daily activity recognition
  via multitask clustering,'' \emph{IEEE Transactions on Image Processing},
  vol.~24, no.~10, pp. 2984--2995, 2015.

\bibitem{RN130}
S.~Chernbumroong, S.~Cang, A.~Atkins, and H.~Yu, ``Elderly activities
  recognition and classification for applications in assisted living,''
  \emph{Expert Systems with Applications}, vol.~40, no.~5, pp. 1662--1674,
  2013.

\bibitem{RN24}
K.-C. Liu, C.-Y. Yen, L.-H. Chang, C.-Y. Hsieh, and C.-T. Chan, ``Wearable
  sensor-based activity recognition for housekeeping task,'' in \emph{IEEE 14th
  International Conference on Wearable and Implantable Body Sensor Networks
  (BSN)}.\hskip 1em plus 0.5em minus 0.4em\relax IEEE, 2017, Conference
  Proceedings, pp. 67--70.

\bibitem{RN29}
L.~Wang, T.~Gu, X.~Tao, and J.~Lu, ``Toward a wearable rfid system for
  real-time activity recognition using radio patterns,'' \emph{IEEE
  Transactions on Mobile Computing}, vol.~16, no.~1, pp. 228--242, 2017.

\bibitem{RN137}
M.~Stikic, T.~Huynh, K.~Van~Laerhoven, and B.~Schiele, ``Adl recognition based
  on the combination of rfid and accelerometer sensing,'' in \emph{2nd
  International Conference on Pervasive Computing Technologies for Healthcare,
  (PervasiveHealth)}.\hskip 1em plus 0.5em minus 0.4em\relax IEEE, 2008,
  Conference Proceedings, pp. 258--263.

\bibitem{RN138}
A.~Hein and T.~Kirste, ``A hybrid approach for recognizing adls and care
  activities using inertial sensors and rfid,'' in \emph{International
  Conference on Universal Access in Human-Computer Interaction}.\hskip 1em plus
  0.5em minus 0.4em\relax Springer, 2009, Conference Proceedings, pp. 178--188.

\bibitem{RN132}
I.~M.~S. Pires, N.~M. Garcia, N.~Pombo, F.~Fl{\'o}rez-Revuelta, E.~Zdravevski,
  and S.~Spinsante, ``Machine learning algorithms for the identification of
  activities of daily living using mobile devices: A comprehensive review,''
  2018.

\bibitem{RN133}
E.~Hoque and J.~Stankovic, ``Aalo: Activity recognition in smart homes using
  active learning in the presence of overlapped activities,'' in \emph{6th
  International Conference on Pervasive Computing Technologies for Healthcare
  (PervasiveHealth)}.\hskip 1em plus 0.5em minus 0.4em\relax IEEE, 2012,
  Conference Proceedings, pp. 139--146.

\bibitem{RN134}
K.~Moriya, E.~Nakagawa, M.~Fujimoto, H.~Suwa, Y.~Arakawa, A.~Kimura, S.~Miki,
  and K.~Yasumoto, ``Daily living activity recognition with echonet lite
  appliances and motion sensors,'' in \emph{IEEE International Conference on
  Pervasive Computing and Communications Workshops (PerCom Workshops)}.\hskip
  1em plus 0.5em minus 0.4em\relax IEEE, 2017, Conference Proceedings, pp.
  437--442.

\bibitem{RN136}
B.~Alsinglawi, Q.~V. Nguyen, U.~Gunawardana, A.~Maeder, and S.~Simoff, ``Rfid
  systems in healthcare settings and activity of daily living in smart homes: A
  review,'' \emph{E-Health Telecommunication Systems and Networks}, vol.~6,
  no.~01, p.~1, 2017.

\bibitem{RN135}
M.~Buettner, R.~Prasad, M.~Philipose, and D.~Wetherall, ``Recognizing daily
  activities with rfid-based sensors,'' in \emph{11th international conference
  on Ubiquitous computing}.\hskip 1em plus 0.5em minus 0.4em\relax ACM, 2009,
  Conference Proceedings, pp. 51--60.

\bibitem{RN139}
P.~Rashidi and A.~Mihailidis, ``A survey on ambient-assisted living tools for
  older adults,'' \emph{IEEE journal of biomedical and health informatics},
  vol.~17, no.~3, pp. 579--590, 2013.

\bibitem{RN140}
A.~Queiros, A.~Dias, A.~G. Silva, and N.~P. Rocha, ``Ambient assisted living
  and health-related outcomes - a systematic literature review,'' in
  \emph{Informatics}, vol.~4.\hskip 1em plus 0.5em minus 0.4em\relax
  Multidisciplinary Digital Publishing Institute, 2017, Conference Proceedings,
  p.~19.

\bibitem{RN141}
G.~Anitha and S.~B. Priya, ``Posture based health monitoring and unusual
  behavior recognition system for elderly using dynamic bayesian network,''
  \emph{Cluster Computing}, pp. 1--8, 2018.

\bibitem{RN142}
C.~Zhu and W.~Sheng, ``Wearable sensor-based hand gesture and daily activity
  recognition for robot-assisted living,'' \emph{IEEE Transactions on Systems,
  Man, and Cybernetics-Part A: Systems and Humans}, vol.~41, no.~3, pp.
  569--573, 2011.

\bibitem{RN117}
H.~Li, C.~Ye, and A.~P. Sample, ``Idsense: A human object interaction detection
  system based on passive uhf rfid,'' in \emph{33rd Annual ACM Conference on
  Human Factors in Computing Systems}.\hskip 1em plus 0.5em minus 0.4em\relax
  ACM, 2015, Conference Proceedings, pp. 2555--2564.

\bibitem{RN143}
Y.~Fouquet, C.~Franco, N.~Vuillerme, and J.~Demongeot, ``Behavioral
  telemonitoring of the elderly at home,'' \emph{International Journal of
  Assistive Robotics and Systems}, vol.~10, pp. 35--49, 2009.

\bibitem{RN144}
R.~Parada, J.~Melia-Segui, M.~Morenza-Cinos, A.~Carreras, and R.~Pous, ``Using
  rfid to detect interactions in ambient assisted living environments,''
  \emph{IEEE Intelligent Systems}, vol.~30, no.~4, pp. 16--22, 2015.

\bibitem{RN145}
M.~A. Soliman and M.~Alrashed, ``An rfid based activity of daily living for
  elderly with alzheimer’s,'' \emph{Internet of Things (IoT) Technologies for
  HealthCare}, p.~54, 2018.

\bibitem{adame2018cuidats}
T.~Adame, A.~Bel, A.~Carreras, J.~Meli{\`a}-Segu{\'\i}, M.~Oliver, and R.~Pous,
  ``Cuidats: An rfid--wsn hybrid monitoring system for smart health care
  environments,'' \emph{Future Generation Computer Systems}, vol.~78, pp.
  602--615, 2018.

\bibitem{RN98}
G.~Ligorio and A.~M. Sabatini, ``A novel kalman filter for human motion
  tracking with an inertial-based dynamic inclinometer,'' \emph{IEEE
  Transactions on Biomedical Engineering}, vol.~62, no.~8, pp. 2033--2043,
  2015.

\bibitem{RN100}
K.~Qian, C.~Wu, Z.~Yang, Y.~Liu, and K.~Jamieson, ``Widar: decimeter-level
  passive tracking via velocity monitoring with commodity wi-fi,'' in
  \emph{18th ACM International Symposium on Mobile Ad Hoc Networking and
  Computing}.\hskip 1em plus 0.5em minus 0.4em\relax ACM, 2017, Conference
  Proceedings, p.~6.

\bibitem{RN101}
X.~Li, D.~Zhang, Q.~Lv, J.~Xiong, S.~Li, Y.~Zhang, and H.~Mei, ``Indotrack:
  Device-free indoor human tracking with commodity wi-fi,'' \emph{Proceedings
  of the ACM on Interactive, Mobile, Wearable and Ubiquitous Technologies},
  vol.~1, no.~3, p.~72, 2017.

\bibitem{RN7}
D.~Zhang, J.~Zhou, M.~Guo, J.~Cao, and T.~Li, ``Tasa: Tag-free activity sensing
  using rfid tag arrays,'' \emph{IEEE Transactions on Parallel and Distributed
  Systems}, vol.~22, no.~4, pp. 558--570, 2011.

\bibitem{RN11}
Y.~Liu, Y.~Zhao, L.~Chen, J.~Pei, and J.~Han, ``Mining frequent trajectory
  patterns for activity monitoring using radio frequency tag arrays,''
  \emph{IEEE Transactions on Parallel and Distributed Systems}, vol.~23,
  no.~11, pp. 2138--2149, 2012.

\bibitem{RN5}
W.~Ruan, L.~Yao, Q.~Z. Sheng, N.~J. Falkner, and X.~Li, ``Tagtrack: Device-free
  localization and tracking using passive rfid tags,'' in \emph{11th
  international conference on mobile and ubiquitous systems: computing,
  networking and services}.\hskip 1em plus 0.5em minus 0.4em\relax ICST
  (Institute for Computer Sciences, Social-Informatics and Telecommunications
  Engineering), 2014, Conference Proceedings, pp. 80--89.

\bibitem{RN103}
L.~Yang, Q.~Lin, X.~Li, T.~Liu, and Y.~Liu, ``See through walls with cots rfid
  system!'' in \emph{21st Annual International Conference on Mobile Computing
  and Networking}.\hskip 1em plus 0.5em minus 0.4em\relax ACM, 2015, Conference
  Proceedings, pp. 487--499.

\bibitem{RN104}
J.~Han, C.~Qian, X.~Wang, D.~Ma, J.~Zhao, W.~Xi, Z.~Jiang, and Z.~Wang,
  ``Twins: Device-free object tracking using passive tags,'' \emph{IEEE/ACM
  Transactions on Networking (TON)}, vol.~24, no.~3, pp. 1605--1617, 2016.

\bibitem{RN99}
W.~Ruan, Q.~Z. Sheng, L.~Yao, T.~Gu, M.~Ruta, and L.~Shangguan, ``Device-free
  indoor localization and tracking through human-object interactions,'' in
  \emph{IEEE 17th International Symposium on World of Wireless, Mobile and
  Multimedia Networks (WoWMoM)}.\hskip 1em plus 0.5em minus 0.4em\relax IEEE,
  2016, Conference Proceedings, pp. 1--9.

\bibitem{RN106}
L.~Chang, X.~Chen, Y.~Wang, D.~Fang, J.~Wang, T.~Xing, and Z.~Tang, ``Fitloc:
  Fine-grained and low-cost device-free localization for multiple targets over
  various areas,'' \emph{IEEE/ACM Transactions on Networking (TON)}, vol.~25,
  no.~4, pp. 1994--2007, 2017.

\bibitem{RN9}
R.~Agrawal, T.~Imielinski, and A.~Swami, ``Mining association rules between
  sets of items in large databases,'' in \emph{Acm sigmod record},
  vol.~22.\hskip 1em plus 0.5em minus 0.4em\relax ACM, 1993, Conference
  Proceedings, pp. 207--216.

\bibitem{RN10}
J.~Han, J.~Pei, and Y.~Yin, ``Mining frequent patterns without candidate
  generation,'' in \emph{ACM sigmod record}, vol.~29.\hskip 1em plus 0.5em
  minus 0.4em\relax ACM, Conference Proceedings, pp. 1--12.

\bibitem{RN107}
A.~N. Ansari, M.~Sedky, N.~Sharma, and A.~Tyagi, ``An internet of things
  approach for motion detection using raspberry pi,'' in \emph{International
  Conference on Intelligent Computing and Internet of Things}.\hskip 1em plus
  0.5em minus 0.4em\relax IEEE, 2015, pp. 131--134.

\bibitem{RN108}
M.~Moghavvemi and L.~C. Seng, ``Pyroelectric infrared sensor for intruder
  detection,'' in \emph{IEEE Region 10 Conference (TENCON)}, vol. 500.\hskip
  1em plus 0.5em minus 0.4em\relax IEEE, 2004, Conference Proceedings, pp.
  656--659.

\bibitem{RN109}
M.~Singh, R.~D. Baruah, and S.~B. Nair, ``A voting-based sensor fusion approach
  for human presence detection,'' in \emph{International Conference on
  Intelligent Human Computer Interaction}.\hskip 1em plus 0.5em minus
  0.4em\relax Springer, 2016, Conference Proceedings, pp. 195--206.

\bibitem{RN110}
J.~Xiao, K.~Wu, Y.~Yi, L.~Wang, and L.~M. Ni, ``Fimd: Fine-grained device-free
  motion detection,'' in \emph{18th International Conference on Parallel and
  Distributed Systems (ICPADS)}.\hskip 1em plus 0.5em minus 0.4em\relax IEEE,
  2012, Conference Proceedings, pp. 229--235.

\bibitem{RN111}
G.~Liu, Y.~Li, D.~Li, X.~Ma, and F.~Li, ``Romd: Robust device-free motion
  detection usin phy layer information,'' in \emph{12th Annual IEEE
  International Conference on Sensing, Communication, and Networking
  (SECON)}.\hskip 1em plus 0.5em minus 0.4em\relax IEEE, 2015, Conference
  Proceedings, pp. 154--156.

\bibitem{RN112}
Y.~Gu, J.~Zhan, Y.~Ji, J.~Li, F.~Ren, and S.~Gao, ``Mosense: An rf-based motion
  detection system via off-the-shelf wifi devices,'' \emph{IEEE Internet of
  Things Journal}, vol.~4, no.~6, pp. 2326--2341, 2017.

\bibitem{RN113}
K.~Zhao, C.~Qian, W.~Xi, J.~Han, X.~Liu, Z.~Jiang, and J.~Zhao, ``Emod:
  Efficient motion detection of device-free objects using passive rfid tags,''
  in \emph{23rd International Conference on Network Protocols (ICNP)}.\hskip
  1em plus 0.5em minus 0.4em\relax IEEE, 2015, Conference Proceedings, pp.
  291--301.

\bibitem{RN115}
Z.~Wang, F.~Xiao, N.~Ye, R.~Wang, and P.~Yang, ``A see-through-wall system for
  device-free human motion sensing based on battery-free rfid,'' \emph{ACM
  Transactions on Embedded Computing Systems (TECS)}, vol.~17, no.~1, p.~6,
  2017.

\bibitem{RN125}
Y.-K. Cheng and R.~Y. Chang, ``Device-free indoor people counting using wi-fi
  channel state information for internet of things,'' in \emph{IEEE Global
  Communications Conference (GLOBECOM)}.\hskip 1em plus 0.5em minus 0.4em\relax
  IEEE, 2017, Conference Proceedings, pp. 1--6.

\bibitem{RN121}
Y.-L. Hou and G.~K. Pang, ``People counting and human detection in a
  challenging situation,'' \emph{IEEE transactions on systems, man, and
  cybernetics-part a: systems and humans}, vol.~41, no.~1, pp. 24--33, 2011.

\bibitem{RN122}
J.~C. S.~J. Junior, S.~R. Musse, and C.~R. Jung, ``Crowd analysis using
  computer vision techniques,'' \emph{IEEE Signal Processing Magazine},
  vol.~27, no.~5, pp. 66--77, 2010.

\bibitem{RN123}
J.~Y. Kuo, G.~D. Fan, and T.~Y. Lai, ``People counting base on head and
  shoulder information,'' in \emph{IEEE International Conference on Knowledge
  Engineering and Applications (ICKEA)}.\hskip 1em plus 0.5em minus 0.4em\relax
  IEEE, 2016, Conference Proceedings, pp. 52--55.

\bibitem{RN124}
H.~Wu, C.~Gao, Y.~Cui, and R.~Wang, ``Multipoint infrared laser-based detection
  and tracking for people counting,'' \emph{Neural Computing and Applications},
  vol.~29, no.~5, pp. 1405--1416, 2018.

\bibitem{RN127}
T.~Teixeira, G.~Dublon, and A.~Savvides, ``A survey of human-sensing: Methods
  for detecting presence, count, location, track, and identity,'' \emph{ACM
  Computing Surveys}, vol.~5, no.~1, pp. 59--69, 2010.

\bibitem{RN126}
H.~Ding, J.~Han, A.~X. Liu, J.~Zhao, P.~Yang, W.~Xi, and Z.~Jiang, ``Human
  object estimation via backscattered radio frequency signal,'' in \emph{IEEE
  Conference on Computer Communications (INFOCOM)}.\hskip 1em plus 0.5em minus
  0.4em\relax IEEE, 2015, Conference Proceedings, pp. 1652--1660.

\bibitem{xie2018multi}
L.~Xie, C.~Wang, A.~X. Liu, J.~Sun, and S.~Lu, ``Multi-touch in the air:
  Concurrent micromovement recognition using rf signals,'' \emph{IEEE/ACM
  Transactions on Networking}, vol.~26, no.~1, pp. 231--244, 2018.

\bibitem{RN116}
L.~Kriara, M.~Alsup, G.~Corbellini, M.~Trotter, J.~D. Griffin, and S.~Mangold,
  ``Rfid shakables: Pairing radio-frequency identification tags with the help
  of gesture recognition,'' in \emph{9th ACM conference on Emerging networking
  experiments and technologies}.\hskip 1em plus 0.5em minus 0.4em\relax ACM,
  2013, Conference Proceedings, pp. 327--332.

\bibitem{RN118}
H.~Li, E.~Brockmeyer, E.~J. Carter, J.~Fromm, S.~E. Hudson, S.~N. Patel, and
  A.~Sample, ``Paperid: A technique for drawing functional battery-free
  wireless interfaces on paper,'' in \emph{Conference on Human Factors in
  Computing Systems (CHI)}.\hskip 1em plus 0.5em minus 0.4em\relax ACM, 2016,
  Conference Proceedings, pp. 5885--5896.

\bibitem{RN119}
S.~Pradhan, E.~Chai, K.~Sundaresan, L.~Qiu, M.~A. Khojastepour, and
  S.~Rangarajan, ``Rio: A pervasive rfid-based touch gesture interface,'' in
  \emph{23rd Annual International Conference on Mobile Computing and
  Networking}.\hskip 1em plus 0.5em minus 0.4em\relax ACM, 2017, Conference
  Proceedings, pp. 261--274.

\bibitem{RN148}
S.~Ranasinghe, F.~Al~Machot, and H.~C. Mayr, ``A review on applications of
  activity recognition systems with regard to performance and evaluation,''
  \emph{International Journal of Distributed Sensor Networks}, vol.~12, no.~8,
  p. 1550147716665520, 2016.

\bibitem{abc}
{JACK A. GOLDSTONE}, ``{The New Population Bomb},''
  \url{https://www.foreignaffairs.com/articles/2010-01-01/new-population-bomb},
  2018, online; accessed 05 July 2018.

\bibitem{RN147}
M.-T. Vo, T.~T. Nghi, V.-S. Tran, L.~Mai, and C.-T. Le, ``Wireless sensor
  network for real time healthcare monitoring: network design and performance
  evaluation simulation,'' in \emph{5th International Conference on Biomedical
  Engineering in Vietnam}.\hskip 1em plus 0.5em minus 0.4em\relax Springer,
  2015, Conference Proceedings, pp. 87--91.

\bibitem{RN149}
B.~Chikhaoui, S.~Wang, and H.~Pigot, ``Activity recognition in smart
  environments: An information retrieval problem,'' in \emph{International
  Conference on Smart Homes and Health Telematics}.\hskip 1em plus 0.5em minus
  0.4em\relax Springer, 2011, Conference Proceedings, pp. 33--40.

\bibitem{RN150}
J.~Cumin, G.~Lefebvre, F.~Ramparany, and J.~L. Crowley, ``Human activity
  recognition using place-based decision fusion in smart homes,'' ser. Modeling
  and Using Context.\hskip 1em plus 0.5em minus 0.4em\relax Springer
  International Publishing, 2017, Conference Proceedings, pp. 137--150.

\bibitem{RN152}
J.~Rafferty, C.~D. Nugent, J.~Liu, and L.~Chen, ``From activity recognition to
  intention recognition for assisted living within smart homes,'' \emph{IEEE
  Transactions on Human-Machine Systems}, vol.~47, no.~3, pp. 368--379, 2017.

\bibitem{RN153}
R.~K. Tripathi, A.~S. Jalal, and S.~C. Agrawal, ``Suspicious human activity
  recognition: a review,'' \emph{Artificial Intelligence Review}, pp. 1--57,
  2017.

\bibitem{blanke2010remember}
U.~Blanke and B.~Schiele, ``Remember and transfer what you have
  learned-recognizing composite activities based on activity spotting,'' in
  \emph{International Symposium on Wearable Computers (ISWC)}.\hskip 1em plus
  0.5em minus 0.4em\relax IEEE, 2010, pp. 1--8.

\bibitem{wang2009sensor}
L.~Wang, T.~Gu, X.~Tao, and J.~Lu, ``Sensor-based human activity recognition in
  a multi-user scenario,'' in \emph{European Conference on Ambient
  Intelligence}.\hskip 1em plus 0.5em minus 0.4em\relax Springer, 2009, pp.
  78--87.

\bibitem{singla2010recognizing}
G.~Singla, D.~J. Cook, and M.~Schmitter-Edgecombe, ``Recognizing independent
  and joint activities among multiple residents in smart environments,''
  \emph{Journal of ambient intelligence and humanized computing}, vol.~1,
  no.~1, pp. 57--63, 2010.

\bibitem{helaoui2011recognizing}
R.~Helaoui, M.~Niepert, and H.~Stuckenschmidt, ``Recognizing interleaved and
  concurrent activities: A statistical-relational approach,'' in \emph{IEEE
  International Conference on Pervasive Computing and Communications
  (PerCom)}.\hskip 1em plus 0.5em minus 0.4em\relax IEEE, 2011, pp. 1--9.

\bibitem{yin2008sensor}
J.~Yin, Q.~Yang, and J.~J. Pan, ``Sensor-based abnormal human-activity
  detection,'' \emph{IEEE Transactions on Knowledge and Data Engineering},
  vol.~20, no.~8, pp. 1082--1090, 2008.

\end{thebibliography}

\end{document}